\documentclass[10pt,twocolumn,letterpaper]{article}

\usepackage[pagenumbers]{cvpr} %

\usepackage{graphicx}
\usepackage{amsmath}
\usepackage{amssymb}
\usepackage{booktabs}

\usepackage[dvipsnames]{xcolor}
\definecolor{cvprblue}{rgb}{0.21,0.49,0.74}
\usepackage[pagebackref,breaklinks,colorlinks]{hyperref}

\usepackage[capitalize]{cleveref}
\crefname{section}{Sec.}{Secs.}
\Crefname{section}{Section}{Sections}
\Crefname{table}{Table}{Tables}
\crefname{table}{Tab.}{Tabs.}
\crefname{algocf}{alg.}{algs.}
\Crefname{algocf}{Algorithm}{Algorithms}

\usepackage[linesnumbered,ruled]{algorithm2e}
\SetKwInOut{Parameters}{Parameters}
\usepackage{makecell}

\usepackage{float}
\usepackage{caption}
\usepackage{subcaption}
\usepackage{stfloats}

\def\confName{WACV}
\def\confYear{2024}

\begin{document}

\title{Multi-Class Segmentation from Aerial Views using Recursive Noise Diffusion}

\author{Benedikt Kolbeinsson\\
Imperial College London\\
{\tt\small bk915@imperial.ac.uk}
\and
Krystian Mikolajczyk\\
Imperial College London\\
{\tt\small k.mikolajczyk@imperial.ac.uk}
}
\maketitle

\begin{abstract}
   Semantic segmentation from aerial views is a crucial task for autonomous drones, as they rely on precise and accurate segmentation to navigate safely and efficiently. However, aerial images present unique challenges such as diverse viewpoints, extreme scale variations, and high scene complexity. In this paper, we propose an end-to-end multi-class semantic segmentation diffusion model that addresses these challenges. We introduce recursive denoising to allow information to propagate through the denoising process, as well as a hierarchical multi-scale approach that complements the diffusion process. Our method achieves promising results on the UAVid dataset and state-of-the-art performance on the Vaihingen Building segmentation benchmark. Being the first iteration of this method, it shows great promise for future improvements. Our code and models are available at: \url{https://github.com/benediktkol/recursive-noise-diffusion}
\end{abstract}

\section{Introduction}\label{sec:intro}
Fully autonomous drones, also known as unmanned aerial vehicles (UAVs), have many socially and economically important applications, such as: infrastructure inspection, agriculture monitoring, search and rescue, disaster management, surveying and delivery services. However, drones need to understand their environment in order to perform these tasks autonomously \cite{floreano2015science}. Semantic segmentation is the task of labeling each pixel within an image to a class (e.g., ``person'' or ``tree'') and is crucial for drones. Segmenting images from aerial views is especially challenging as they include diverse view-points, extreme scale variation and high scene complexity \cite{lyu2020uavid}. 

Conventional methods typically utilize a convolutional neural network (CNN) with an encoder-decoder structure, such as U-Net \cite{ronneberger2015u}. A limited receptive field will struggle with large scale variations, for example detecting large objects benefits from more global features while smaller objects are often better predicted using larger image sizes \cite{tao2020hierarchical}. To overcome these limitations it is common to use multi-scale features \cite{farabet2012learning, mostajabi2015feedforward}. Combining scales by averaging the output from different scales at inference works but is improved when the scale combination is learned \cite{chen2016attention, tao2020hierarchical}.

\begin{figure}[t]
    \centering
    \includegraphics[width=0.8\linewidth]{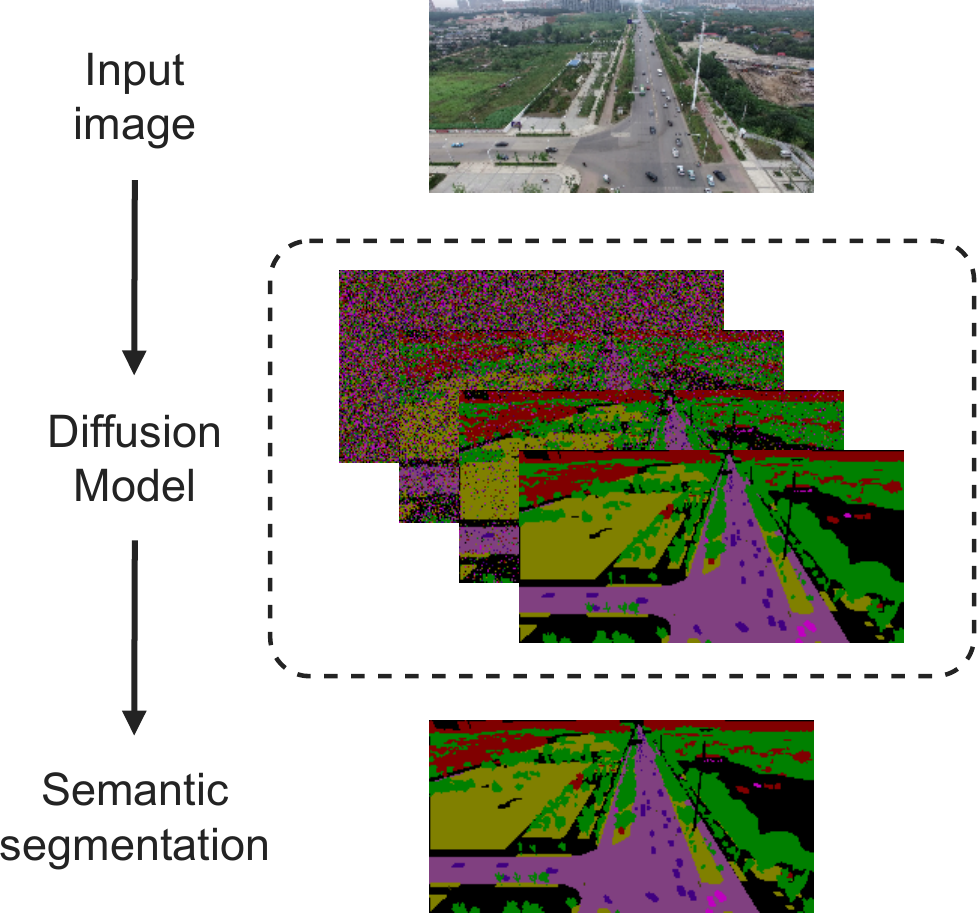}
    \caption{\textbf{A high level illustration of the recursive diffusion concept.} The diffusion model is conditioned on the input image as well as the previous segmentation prediction at various scales, before returning the final semantic segmentation map.}
    \label{fig:intro}
\end{figure}

To address this problem, we propose a hierarchical multi-scale diffusion model, which naturally learns how to combine multi-scale predictions (shown in \Cref{fig:intro}). This is made possible as we introduce a novel training method called \textit{recursive denoising}, which allows a diffusion model to learn to use additional information during inference. We discuss the details of our method in \Cref{sec:method} and demonstrate competitive results on UAVid~\cite{lyu2020uavid} and state-of-the-art results on Vaihingen Buildings~\cite{cramer2010dgpf} in \Cref{sec:experiments}.

Our main contributions are:
\begin{itemize}
    \item We introduce \textit{recursive denoising}, which allows information to propagate through the denoising process, along with a hierarchical multi-scale approach.
    
    \item We implement a diffusion model for multi-class segmentation using \textit{recursive denoising}.
    
    \item We achieve promising results on UAVid and state-of-the-art results on Vaihingen Buildings. 
\end{itemize}

\section{Related Work}
\paragraph{Semantic Segmentation}

Over the years, numerous methods have been proposed to tackle semantic segmentation, ranging from classical methods based on hand-crafted features to deep learning-based approaches. Long \etal~\cite{long2015fully} introduced fully convolutional neural networks (CNNs), which have subsequently emerged as the dominant paradigm for semantic segmentation \cite{chen2014semantic, chen2017deeplab, peng2017large}. The success of CNNs can be attributed to their ability to automatically learn meaningful representations from raw data, as well as their ability to capture context information through the use of large receptive fields \cite{luo2016understanding}. In addition, many works have explored the use of different architectures and training strategies to improve the accuracy and efficiency of semantic segmentation models. For example, to increase the receptive field of a network, Fisher Yu and Vladlen Koltun~\cite{yu2015multi} used dilated convolutions,  later improved by Yu \etal~\cite{yu2017dilated} with dilated residual networks. 
PSPNet~\cite{zhao2017pyramid} and DeepLab~\cite{chen2018encoder} use pyramid pooling, while Tao \etal~\cite{tao2020hierarchical} and Chen \etal~\cite{chen2016attention} use multi-scale attention to increase the network's receptive field. Additionally, some approaches have used attention mechanisms~\cite{vaswani2017attention} to selectively focus on informative regions of the image \cite{zheng2021rethinking, xie2021segmenting, Strudel_2021_ICCV, xie2021segformer}. 

For aerial views, Lyu \etal~\cite{lyu2020uavid} propose a multi-scale-dilation network to combat the large scale variations. The use of attention is more common, where Wang \etal~\cite{wang2021transformer} develop a bilateral awareness network and Yi \etal~\cite{yi2023uavformer} use a composite encoder from which multi-scale features are extracted. 
Yang \etal~\cite{yang2021real} use a dual branch approach, one branch for high resolution spacial details and the other for global aggregation. While, Wang \etal~\cite{wang2022unetformer} explore using attention in the decoder stage for efficiency. A dual branch encoder with different scales and a cross-scale attention module is proposed by Ding \etal~\cite{ding2022dense} with state-of-the-art results.

Overall, semantic segmentation remains an active and challenging research area with numerous opportunities for further exploration and improvement.

\paragraph{Diffusion}
Denoising diffusion probabilistic models (DDPM), also known simply as diffusion models, are a type of generative models. Initially proposed by Sohl-Dickstein \etal~\cite{sohl2015deep}, they have become state-of-the-art as they have been shown to outperform GANs \cite{dhariwal2021diffusion}. Thus, they have recently become exceedingly popular for image generation \cite{croitoru2022diffusion, saharia2022photorealistic, rombach2022high, ramesh2022hierarchical}. In addition, diffusion models are starting to be applied to a wide range of problems including object detection \cite{chen2022diffusiondet}, video editing \cite{molad2023dreamix} and music generation \cite{agostinelli2023musiclm}.

\paragraph{Segmentation using generative methods} 
Generative adversarial networks (GANs)~\cite{NIPS2014_goodfellow_generative} have been used to generate synthetic data to train segmentation models \cite{benjdira2019unsupervised, toldo2020unsupervised}. GANs are also used to predict semantic segmentation directly \cite{luc2016semantic, zhang2018seggan, souly2017semi}.

Baranchuk \etal~\cite{baranchuk2021label} use intermediate representations from a pre-trained diffusion model to perform semantic segmentation. Amit \etal~\cite{amit2021segdiff} train a diffusion model to perform segmentation but only for binary classification. Similarly, Wu \etal~\cite{wu2022medsegdiff} use a diffusion model for binary segmentation for medical images. Both of these methods are inherently limited to binary classification. We propose modifications to diffusion models which allow us to perform full multi-class semantic segmentation. We show our method achieves state-of-the-art results on the Vaihingen building dataset and promising results on UAVid in \Cref{sec:experiments}.

\begin{figure*}[t]
    \centering
    \includegraphics[width=1\linewidth]{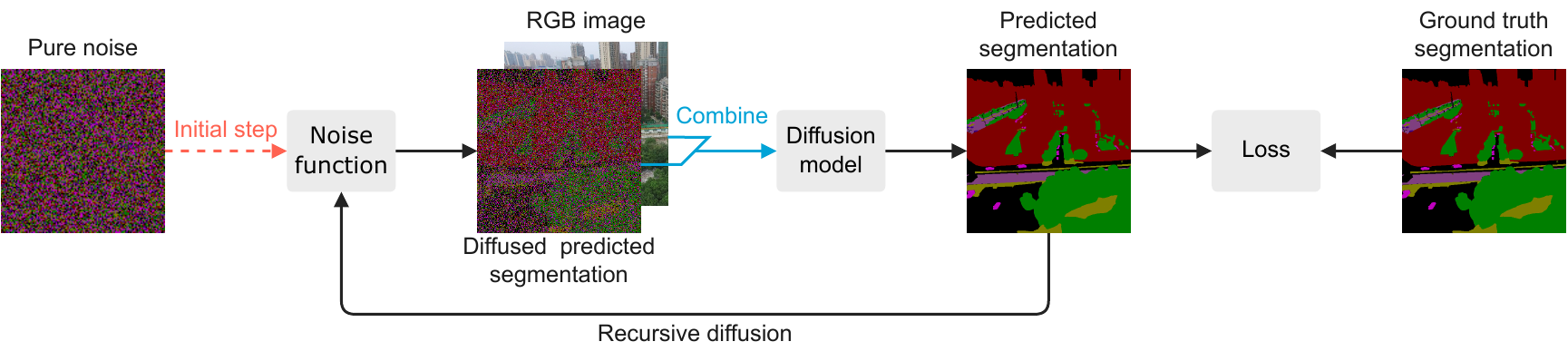}
    \caption{\textbf{Overview of the \textit{recursive noise diffusion} process.} The noise function diffuses the previous predicted segmentation. The model denoises the diffused segmentation given a conditioning RGB image. Finally, the denoised predicted segmentation is compared to the ground truth. The segmentation is initialized as pure noise. Notably, the ground truth segmentation is never used as part of the input to the model. This process is agnostic to the choice of noise function, diffusion model and loss.}
    \label{fig:rnd_diagram}
\end{figure*}

\paragraph{Diffusion Theory}
Diffusion models consist of two processes, first the \textit{diffusion process} or \textit{forward process} \(q\), a Markov chain which gradually adds Gaussian noise to the data. Given a data distribution \(\mathbf{x}_0 \sim q(\mathbf{x}_0)\), we define \(q\) as:

\begin{equation}
    q(\mathbf{x}_{1:T}|\mathbf{x}_{0}) = \prod_{t=1}^{T} q(\mathbf{x}_{t}|\mathbf{x}_{t-1})
\end{equation}

\begin{equation}
    q(\mathbf{x}_{t}|\mathbf{x}_{t-1}) = \mathcal{N}\left(\mathbf{x}_{t};\sqrt{1-\beta_{t}\mathbf{x}_{t-1}},\beta_{t}\mathbf{I}\right)
\end{equation}

Where \(\beta_t \in (0,1)\) is the noise schedule, \(\mathbf{x}_{1},\dots,\mathbf{x}_{T}\) are latent variables and \(T\) is the number of time steps. The latent \(\mathbf{x}_T\) is approximately an isotropic Gaussian distribution, given a sufficiently large \(T\).

The second process, called the \textit{denoising process} or \textit{reverse process} \(p\), is also defined as a Markov chain starting at \(p_{\theta}(\mathbf{x}_{T}) = \mathcal{N}(\mathbf{x}_{T};\mathbf{0},\mathbf{I})\) but learns to gradually remove Gaussian noise:

\begin{equation}
    p_{\theta}(\mathbf{x}_{0:T}) = p_{\theta}(\mathbf{x}_T) \prod_{t=1}^{T} p_{\theta}(\mathbf{x}_{t-1}|\mathbf{x}_{t})
\end{equation}

\begin{equation}
    p_{\theta}(\mathbf{x}_{t-1}|\mathbf{x}_{t}) = \mathcal{N}\left(\mathbf{x}_{t-1};\mathbf{\mu}_\theta(\mathbf{x}_{t},t),\mathbf{\Sigma}_{\theta}(\mathbf{x}_{t},t)\right)
\end{equation}

Interestingly, given \(\alpha_{t} = 1-\beta_{t}\) and \(\bar{\alpha}_{t} = \prod_{s=0}^{t}\alpha_{s}\), sampling \(\mathbf{x}_{t}\) at arbitrary timestep \(t\) is achieved with:

\begin{equation}
     q(\mathbf{x}_{t}|\mathbf{x}_{0}) = \mathcal{N}\left(\mathbf{x}_{t};\sqrt{\bar{\alpha}_{t}\mathbf{x}_{0}},(1-\bar\alpha_{t})\mathbf{I}\right)
\end{equation}

Further details on the theory behind diffusion models may be found in~\cite{ho2020denoising, nichol2021improved}. However, Bansal \etal~\cite{bansal2022cold} call this theory into question. They show diffusion models can be trained without Gaussian noise and even with deterministic image degradation. We are inspired by their work, and thus we propose to use a diffusion model to predict semantic segmentation maps. Converting diffusion models from typical generative tasks, to a predictive task, involves modifying the \textit{diffusion process} and the \textit{denoising process}. We discuss the details in \Cref{sec:method}.

\section{Recursive Noise Diffusion}\label{sec:method}
Given an aerial RGB image \(\mathbf{x} \in  \mathbb{R}^{W \times H \times 3}\), our goal is to predict a semantic segmentation map \(\mathbf{s} \in \mathbb{R}^{W \times H \times d_{\textit{classes}}}\), with corresponding class labels for each pixel. Where \(W, H\) are the width and height of the image, and \(d_{\textit{classes}}\) is the total number of classes.

\subsection{Multi-class diffusion}
We modify the diffusion process to better suit the problem of predicting semantic segmentation. Given an image \(\mathbf{x} \in \mathbf{X}\), the corresponding one-hot encoded segmentation map \(\mathbf{s}_{0} \in (\mathbf{S}|\mathbf{x})\) and time step (noise level) \(t \in [0,T]\), we define the forward noising process \(q\), which adds Gaussian noise with variance \(\beta_t \in (0,1)\), as follows:
\begin{equation}
    q(\mathbf{s}_{t} | \mathbf{s}_{t-1}) = \mathbf{s}_{t-1} + \mathcal{N}(0, \beta_{t}\mathbf{I})
\end{equation}
We define the noise schedule \(\beta_{t}\), as:
\begin{equation}
    \beta_{t} = \frac{t}{T}
\end{equation}
Trivially, the total added noise can be written as:
\begin{equation}
    \mathbf{\epsilon}_{t} = \mathbf{s}_{t} - \mathbf{s}_{0}
\end{equation}
We approximate the reverse process (denoising) using a neural network, \(\epsilon_{\theta}\). Following Ho \etal~\cite{ho2020denoising}, \(\epsilon_{\theta}\) predicts \(\mathbf{\epsilon}_{t}\), meaning:
\begin{equation}
    \mathbf{\epsilon}_{t} \approx \epsilon_{\theta}(\mathbf{s}_{t}, \mathbf{x}, t)
\end{equation} 
Thus, we can predict the segmentation map at any arbitrary time step \(t\), as follows: 
\begin{equation}\label{eq:predict_segmentation}
    \mathbf{s}_{0} \approx \mathbf{s}_{t} - \epsilon_{\theta}(\mathbf{s}_{t}, \mathbf{x}, t)
\end{equation} 
For training, we use the mean squared error (MSE) of the predicted noise as our loss function:
\begin{equation}\label{eq:mse_loss}
    L_{\text{mse}} = E_{\mathbf{s}_{0},\mathbf{x}, t, \mathbf{\epsilon}_{t}}\left[ \| \mathbf{\epsilon}_{t} - \mathbf{\epsilon}_{\theta}(\mathbf{s}_{t}, \mathbf{x}, t)\|^2 \right]
\end{equation}

\subsection{Recursive denoising}
When we train a model on an arbitrary step of the noising process, as is most common \cite{ho2020denoising,nichol2021improved}, we notice the model quickly overfits on the training data and does not generalize. We believe the binary labels allow for a trivial denoising strategy to be learned, e.g. rounding logits. This causes two issues, first, the model does not fully utilize the conditional image but rather simply uses the noisy segmentation. Second, during testing, the model is too dependent on the initial steps in the denoising process. To solve this, we propose to train with \textit{recursive denoising}.

Training with \textit{recursive denoising} involves progressing through each time step \(t\) from \(T\) to \(1\), recursively (as the name suggests), which allows a portion of the predicted error to propagate.
\Cref{fig:rnd_diagram} shows an overview of the \textit{recursive noise diffusion} method. The first step of the approach involves the use of pure noise. The subsequent stage utilizes a noise function to diffuse the previous predicted segmentation. Following this, a diffusion model is employed to denoise the diffused segmentation, leveraging a conditioning RGB image. Finally, the denoised predicted segmentation is compared to the ground truth. Importantly, it is worth noting that the ground truth segmentation is never utilized as a component of the input to the model. \Cref{alg:recursive_denoising} outlines the steps involved in processing a single training sample. Our recursive denoising approach can be conceptualized as emulating the testing process, which necessitates advancing through each time step.

An essential characteristic of our proposed method is its agnosticism to the selection of the noise function, diffusion model, and loss function. This feature enables the flexibility to choose the most suitable components for a specific task or dataset. To illustrate this capability, we demonstrate the use of a distinct noise function, diffusion model, and loss function in \Cref{sec:multi-class-segmentation} compared to \Cref{sec:binary-segmentation} of our experiments. Such flexibility provides greater versatility and adaptability to the proposed approach, enabling it to be widely applicable in a range of tasks and settings.

\textit{Recursive denoising} can also serve as a means of transmitting supplementary information, as elaborated upon in \Cref{sec:method_multi-scale}.

\begin{algorithm}[ht]
\DontPrintSemicolon
  \KwIn{$\mathbf{x} \in \mathbb{R}^{W \times H \times 3}$, RGB image}
  \KwIn{$\mathbf{\bar{s}} \in \mathbb{R}^{W \times H \times \textit{classes}}$, segmentation labels}
  \Parameters{$T \in \mathbb{Z}^{1}$, number of time steps}
  $\mathbf{\hat{s}}_{T} \sim \mathcal{N}(\mathbf{0},\mathbf{I})$\;
  \For{$t=T,...,1$}{
    $\mathbf{z}_{t} \sim \mathcal{N}(\mathbf{0},\mathbf{I})$\;
    $\mathbf{s}^{\prime}_{t} \gets \mathbf{\hat{s}}_{t} + \mathbf{z}_{t} \times \frac{t}{T} $ \tcp*{diffuse}
    $\mathbf{\hat{s}}_{t-1} \gets \mathbf{s}^{\prime}_{t} - \epsilon_{\theta}(\mathbf{s}^{\prime}_{t}, \mathbf{x}, t)$ \tcp*{denoise}
    $l \gets \lVert \epsilon_{\theta} (\mathbf{s}^{\prime}_{t}, \mathbf{x}, t) - (\mathbf{s}^{\prime}_{t} - \mathbf{\bar{s}}) \rVert ^{2}$\;
    $\text{Update } \epsilon_{\theta} \text{ w.r.t. } l$
  }
\caption{Training with recursive denoising}
\label{alg:recursive_denoising}
\end{algorithm}

\subsection{Hierarchical multi-scale diffusion}\label{sec:method_multi-scale}
\begin{figure*}[t]
    \centering
    \includegraphics[width=0.9\linewidth]{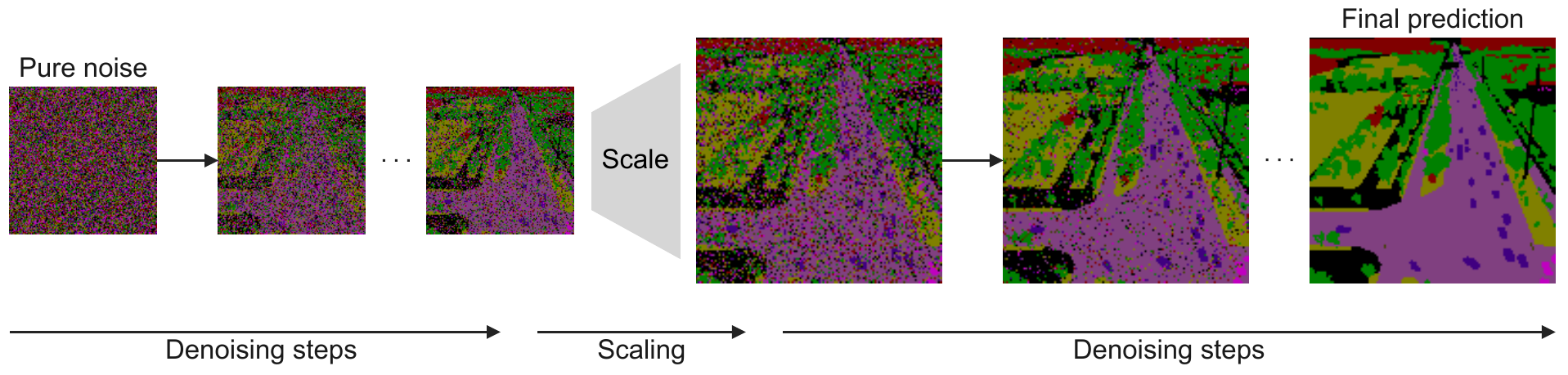}
    \caption{\textbf{The hierarchical multi-scale process.} A down-scaled input is first denoised for half the time steps before up-scaling to the original resolution (bilinear intorpolation) for the remaining time steps. It can be noted that large structures appear first while finer detail appear later.}
    \label{fig:multi_scales}
\end{figure*}

Performing segmentation at multiple image scales can improve the prediction \cite{ding2022dense}. We can exploit the propagation artifact from \textit{recursive denoising} using a hierarchical multi-scale denoising process, shown in \Cref{fig:multi_scales}. The assumption underlying our approach is that the model can extract valuable information from the previous noisy segmentation to improve the accuracy of the next prediction. Leveraging our recursive denoising approach, this assumption can be exploited by enabling the model to make predictions on a smaller (scaled down) image first, thereby capturing larger objects and contextual information, followed by predictions on a higher resolution image to capture finer details and smaller objects.

We define a simple linear scaling schedule to determine when to change scales. Our scaling schedule evenly divides the time steps for each scale, starting with the smallest scale. Our approach is not restricted solely to this particular scaling schedule. In \Cref{sec:binary-segmentation}, we investigate a variation of this scaling approach.

\section{Experiments}\label{sec:experiments}
We demonstrate the versatility of \textit{recursive diffusion} in this section through two experiments: multi-class segmentation and binary segmentation. Each experiment uses a distinct combination of model, noise function, and loss. Experimental settings and results for each experiment are presented separately.

\subsection{Multi-class segmentation}\label{sec:multi-class-segmentation}
In this section, we demonstrate the effectiveness of our method for multi-class segmentation for aerial views. Multi-class segmentation refers to the process of labeling each pixel in an image into more than two distinct classes. This is in contrast to binary segmentation, which involves the separation of pixels in an image into two classes only. Multi-class segmentation is generally more challenging than binary segmentation, as it requires the accurate identification and labeling of multiple objects or regions within an image, each belonging to a distinct class.

\paragraph{Data}
We use UAVid~\cite{lyu2020uavid}, a specialized drone dataset, containing a total of 420 (4K resolution) images from aerial views. The data is split into sets of 200, 70, and 150 images for training, validation and testing, respectively. There are eight classes; \textit{Building}, \textit{Road}, \textit{Tree}, \textit{Low Vegetation}, \textit{Moving Car},  \textit{Static Car}, \textit{Human} and \textit{Clutter}.

\paragraph{Metrics}
We report the Intersection over Union (IoU), also known as the Jaccard Index, for individual classes as well as the mean Intersection over Union (mIoU), averaged over all classes. These metrics are often expressed as percentages.

\begin{equation}
    IoU = \frac{TP}{TP+FP+FN}
\end{equation}

\begin{equation}
    mIoU = \frac{1}{N}\sum_{i=1}^{N}\frac{TP_{i}}{TP_{i}+FP_{i}+FN_{i}}
\end{equation}

Where \textit{TP}, \textit{FP} and \textit{FN} are the true positive, false positive and false negative between the predicted and ground truth class labels, respectively. \textit{N} is the total number of classes in the dataset. We also report the F1-score.

\begin{figure*}[t]
    \centering
    \includegraphics[width=0.78\linewidth]{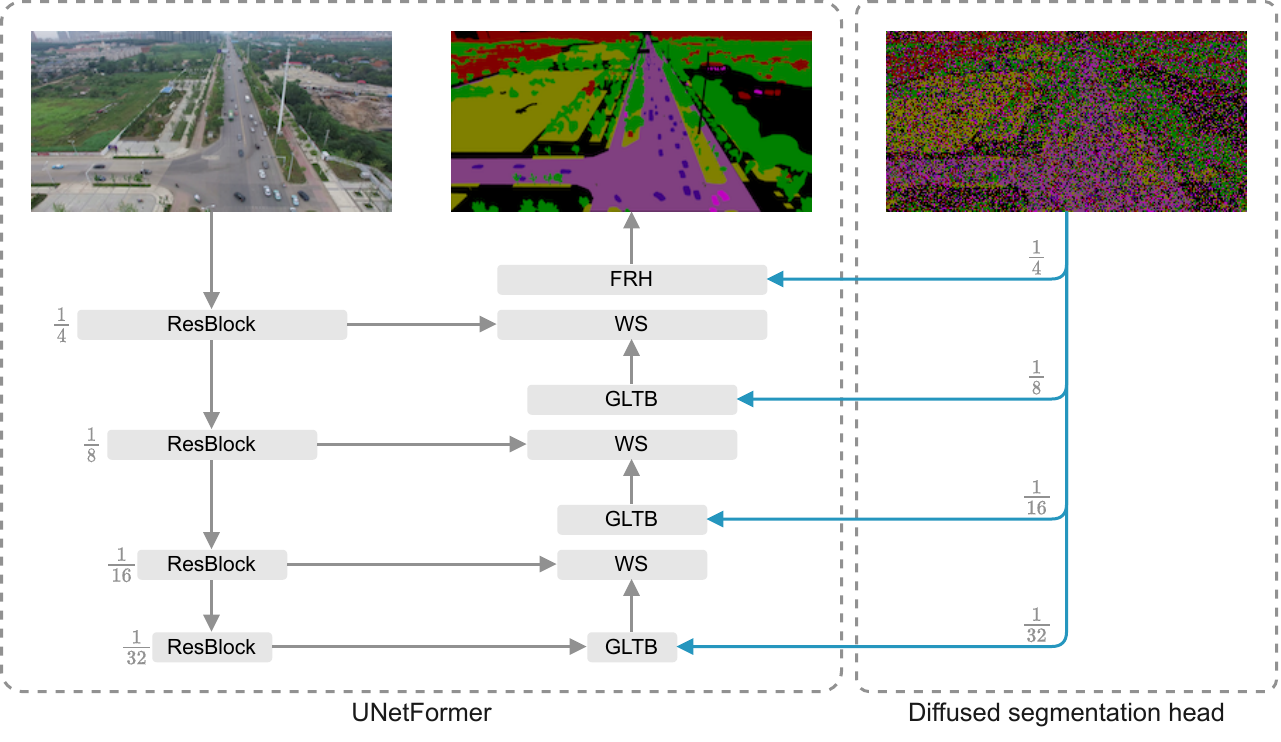}
    \caption{\textbf{WNetFormer model architecture.} Converting UNetFormer to a diffusion model. UNetFormer consists of global-local transformer blocks (GLTB), weighted sums (WS) and a feature refinement head (FRH). The diffused segmentation head consists of down-sampling (bilinear interpolation) and concatenation. The time step is concatenated, channel-wise, to the diffused segmentation.}
    \label{fig:architecture}
\end{figure*}

\paragraph{Architecture}
As the \textit{recursive denoising} approach is independent of the choice of model, it provides us with the flexibility to choose an efficient and effective model for multi-class segmentation of aerial views, such as UNetFormer~\cite{wang2022unetformer}. UNetFormer can be conveniently transformed into a diffusion model by incorporating the prior segmentation as an extra input, which requires only minor modifications to the original architecture. The architectural changes are illustrated in Figure \ref{fig:architecture}. Specifically, the previous segmentation input is scaled (using bilinear interpolation) and concatenated to minimize the introduction of additional network complexity. The resulting WNetFormer, which is a variant of UNetFormer with these modifications, has only a marginal increase in parameters compared to the original UNetFormer, with 11.8 M parameters versus 11.7 M parameters.

\paragraph{Loss}
To demonstrate the versatility of our approach, we employ a loss function similar to that of UNetFormer. UNetFormer incorporates both a primary loss and an auxiliary loss, with the latter being applied to an auxiliary head that integrates features from the global-local transformer block (GLTB). Specifically, the auxiliary loss takes the form of a cross entropy loss. It is noteworthy that we have retained this aspect of the UNetFormer methodology and have not made any modifications to the auxiliary loss function. However, we simplify UNetFormer's primary loss as we only use the cross entropy portion. In addition, we target the added noise instead of directly predicting the segmentation, as discussed in \Cref{sec:method}.

\paragraph{Noise function}
As previously mentioned, \textit{recursive noise diffusion} does not impose any restrictions on the noise function. However, care must be taken when choosing a noise function as the one-hot encoded binary labels for the classes is inherently robust to low levels of noise. We propose a Softmax-Average noise function which is both effective and controllable. Using the same notation as in \Cref{sec:method}, the Softmax-Average noise function is defined per pixel as:
\begin{equation}
    \mathbf{s}^{\prime}_{t,\mathbf{\hat{s}}} = softmax\left( (1 - \lambda_{noise,t}) \times \mathbf{\hat{s}} + \lambda_{noise,t} \times \mathbf{z} \right)
\end{equation}
Where \(\mathbf{z}\) is a softmaxed vector of random noise where each element is sampled from the normal distribution, and
\begin{equation}
    \lambda_{noise,t} = \frac{T-t}{T}
\end{equation}

\paragraph{Training}
Diffusion models tend to use a large number of time steps, in the range of hundreds or thousands \cite{nichol2021improved}. However, we find we can train with far fewer time steps (as low as 2), which directly translates to reduced inference time. However, performance is better when trained with more time steps as can be seen in \Cref{tab:ablation_training_timesteps}, and we train with 128 time steps.
We augment the training data with random (50\%) horizontal flips and adjust the brightness, contrast, saturation and hue. We train on a single NVIDIA GeForce RTX 2080 Ti for 100 epochs with a batch size of 4. We use the AdamW~\cite{loshchilov2017decoupled} optimizer with learning rate of \(6 \times 10^{-4}\) with a cosine annealing schedule and weight decay of \(0.01\).

\begin{table*}[t]
  \centering
  \begin{tabular}{@{}lccccccccc@{}}
    \toprule
    Method & Building & Road & Tree & \makecell{Low \\ Vegetation} & \makecell{Moving \\ Car} & \makecell{Static \\ Car} & Human & Clutter & mIoU\\
    \midrule

    UAVFormer \cite{yi2023uavformer} & 81.5 & 67.1 & 76.2 & 48.5 & 62.2 & 28.8 & 12.5 & 48.8 & 53.2 \\
    CANet \cite{yang2021real} & 86.6 & 62.1 & 79.3 & 78.1 & 47.8 & 68.3 & 19.9 & 66.0 & 63.5 \\
    BANet \cite{wang2021transformer} & 85.4 & 80.7 & 78.9 & 62.1 & 52.8 & 69.3 & 21.0 & 66.6 & 64.6 \\
    UNetFormer \cite{wang2022unetformer} & 87.4 & 81.5 & 80.2 & 63.5 & 73.6 & 56.4 & 31.0 & 68.4 & 67.8 \\
    DCDNet \cite{ding2022dense} & 90.6 & 83.6 & 82.2 & 66.5 & 77.7 & 74.7 & 31.7 & 72.1 & 72.4 \\
    Ours & 87.7 & 80.1 & 79.9 & 63.5 & 71.2 & 60.1 & 26.3 & 68.3 & 67.1 \\
    
    \bottomrule
  \end{tabular}
  \caption{\textbf{Comparison of different methods on the UAVid~\cite{lyu2020uavid} test dataset.} Each column represents IoU per respective class with the right-most column being the mean IoU.}
  \label{tab:uavid}
\end{table*}

\begin{table}
  \centering
  \begin{tabular}{@{}ccc@{}}
    \toprule
    \# of training time steps & mIoU & F1-score \\
    \midrule
    2 & 66.9 & 79.2 \\
    4 & 67.0 & 79.6 \\
    8 & 66.8 & 78.9 \\
    16 & 70.0 & 81.7 \\
    32 & 70.8 & 82.3 \\
    64 & 71.2 & 82.5 \\
    128 & \textbf{71.3} & \textbf{82.6} \\
    
    \bottomrule
  \end{tabular}
  
  \caption{\textbf{The impact of varying the number of time steps during training, on UAVid validation.} More time steps correspond to higher mIoU. Best results are highlighted in bold.}
  
  \label{tab:ablation_training_timesteps}
\end{table}

\begin{table}
  \centering
  \begin{tabular}{@{}ccc@{}}
    \toprule
    \# of inference time steps & mIoU & F1-score \\
    \midrule
    128 & \textbf{71.34} & \textbf{82.59} \\
    64 & 71.18 & 82.49 \\
    32 & 70.64 & 82.14 \\
    16 & 68.71 & 80.84 \\
    8 & 61.57 & 75.70 \\
    4 & 47.02 & 62.74 \\
    2 & 35.05 & 48.61 \\
    
    \bottomrule
  \end{tabular}
  \caption{\textbf{The impact of varying inference time steps.} The effects of different time steps during inference using a model trained with 128 time steps. This approach allows for efficiency improvements while maintaining high performance. Best results are highlighted in bold.}
  \label{tab:ablation_timesteps_inference}
\end{table}

\begin{table}[t]
  \centering
  \begin{tabular}{@{}lcc@{}}
    \toprule
    Description & \# of scales & \(\Delta\) mIoU \\
    \midrule
    Recursive + Scales & 4 & +3.35\% \\
    Recursive + Scales & 3 & \textbf{+6.57\%} \\
    Recursive + Scales & 2 & +5.40\% \\
    Recursive & 1 & - \\
    
    \bottomrule
  \end{tabular}
  \caption{\textbf{The impact of our hierarchical multi-scale approach.} We present performance results in terms of relative mIoU improvement compared to our recursive method without multiple scales on the UAVid dataset. We progressively introduce additional scales and not excessively small scales contain limited additional information. Best results are highlighted in bold.}

  \label{tab:ablation_multi_scale}
\end{table}

\paragraph{Results}
Our proposed \textit{recursive denoising} method achieves promising results, shown in \Cref{tab:uavid}. The method outperforms other recently proposed approaches including BANet \cite{wang2021transformer} and UAVFormer \cite{yi2023uavformer}. Qualitative experiments are shown in \Cref{appendix:qualitative_experiments}. As this is the first iteration of this method there is scope for improvement through optimization of hyperparameters that were not fully explored due to limited computational resources. The Human class is particularly challenging, as it is both underrepresented and each object is small and isolated in relation of the entire scene.
We investigate the effects of varying the number of time steps during training. As presented in \Cref{tab:ablation_training_timesteps}, a discernible trend emerges, indicating that higher numbers of training time steps yield higher segmentation scores.

In order to enhance the efficiency of our method, we further examine the impact of skipping time steps during inference, as illustrated in \Cref{tab:ablation_timesteps_inference}. This is made feasible through the utilization of \Cref{eq:predict_segmentation}. Our findings reveal that a reduction of up to 50\% in the number of time steps can be implemented with minimal effect on performance. However, beyond this threshold, a sharp decline in performance is observed. Therefore, our results demonstrate that our method can be trained with a high number of time steps while simultaneously enhancing its efficiency during inference, without any significant loss of performance.

In addition, we verify the effects of our hierarchical multi-scale approach and demonstrate its direct benefits when used with \textit{recursive denoising} in an ablation study shown in \Cref{tab:ablation_multi_scale}.

\begin{figure*}[t]
     \centering
     \begin{subfigure}[b]{0.12\textwidth}
         \centering
         \includegraphics[width=\textwidth]{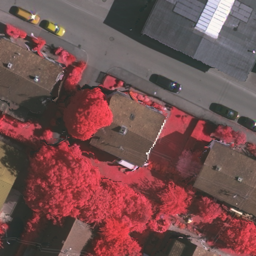}
     \end{subfigure}
     \hfill
     \begin{subfigure}[b]{0.12\textwidth}
         \centering
         \includegraphics[width=\textwidth]{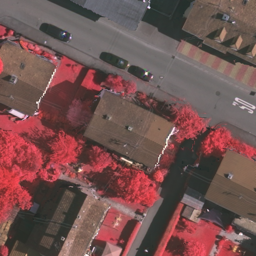}
     \end{subfigure}
     \hfill
     \begin{subfigure}[b]{0.12\textwidth}
         \centering
         \includegraphics[width=\textwidth]{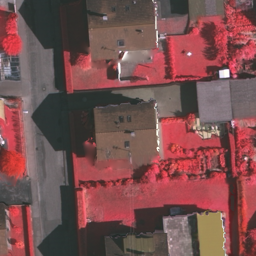}
     \end{subfigure}
     \hfill
     \begin{subfigure}[b]{0.12\textwidth}
         \centering
         \includegraphics[width=\textwidth]{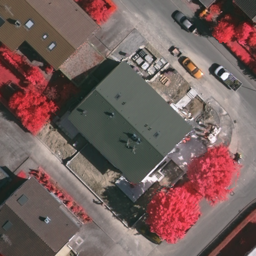}
     \end{subfigure}
     \hfill
     \begin{subfigure}[b]{0.12\textwidth}
         \centering
         \includegraphics[width=\textwidth]{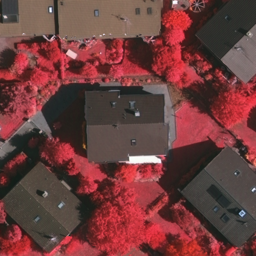}
     \end{subfigure}
     \hfill
     \begin{subfigure}[b]{0.12\textwidth}
         \centering
         \includegraphics[width=\textwidth]{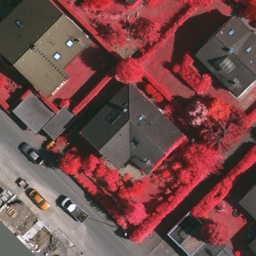}
     \end{subfigure}
     \hfill
     \begin{subfigure}[b]{0.12\textwidth}
         \centering
         \includegraphics[width=\textwidth]{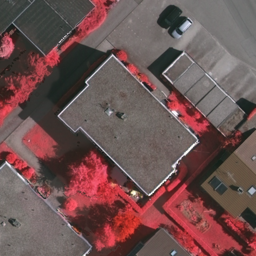}
     \end{subfigure}
     \hfill
     \begin{subfigure}[b]{0.12\textwidth}
         \centering
         \includegraphics[width=\textwidth]{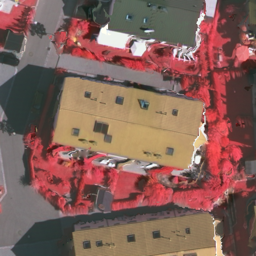}
     \end{subfigure}

     \begin{subfigure}[b]{0.12\textwidth}
         \centering
         \includegraphics[width=\textwidth]{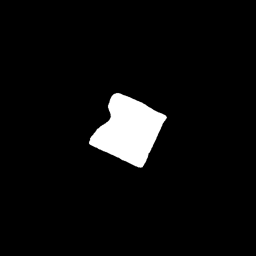}
     \end{subfigure}
     \hfill
     \begin{subfigure}[b]{0.12\textwidth}
         \centering
         \includegraphics[width=\textwidth]{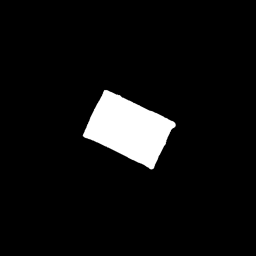}
     \end{subfigure}
     \hfill
     \begin{subfigure}[b]{0.12\textwidth}
         \centering
         \includegraphics[width=\textwidth]{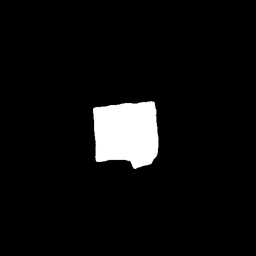}
     \end{subfigure}
     \hfill
     \begin{subfigure}[b]{0.12\textwidth}
         \centering
         \includegraphics[width=\textwidth]{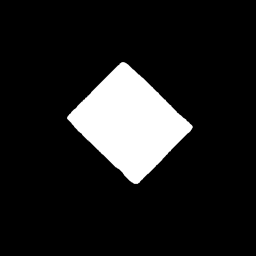}
     \end{subfigure}
     \hfill
     \begin{subfigure}[b]{0.12\textwidth}
         \centering
         \includegraphics[width=\textwidth]{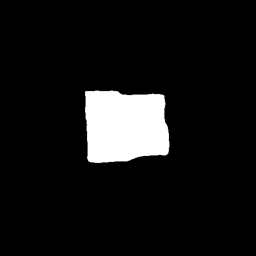}
     \end{subfigure}
     \hfill
     \begin{subfigure}[b]{0.12\textwidth}
         \centering
         \includegraphics[width=\textwidth]{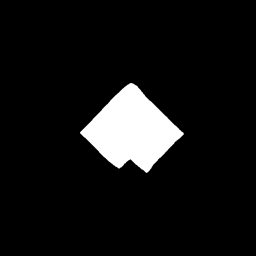}
     \end{subfigure}
     \hfill
     \begin{subfigure}[b]{0.12\textwidth}
         \centering
         \includegraphics[width=\textwidth]{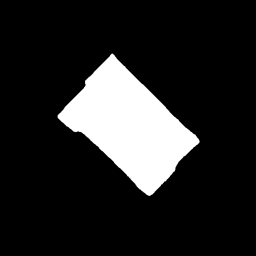}
     \end{subfigure}
     \hfill
     \begin{subfigure}[b]{0.12\textwidth}
         \centering
         \includegraphics[width=\textwidth]{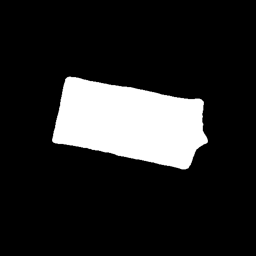}
     \end{subfigure}

     \begin{subfigure}[b]{0.12\textwidth}
         \centering
         \includegraphics[width=\textwidth]{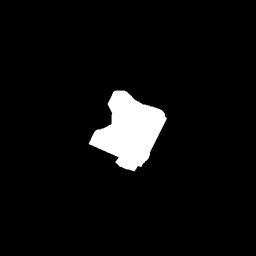}
     \end{subfigure}
     \hfill
     \begin{subfigure}[b]{0.12\textwidth}
         \centering
         \includegraphics[width=\textwidth]{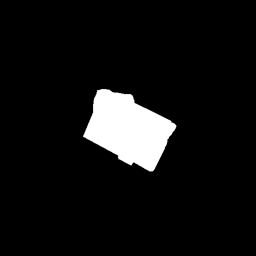}
     \end{subfigure}
     \hfill
     \begin{subfigure}[b]{0.12\textwidth}
         \centering
         \includegraphics[width=\textwidth]{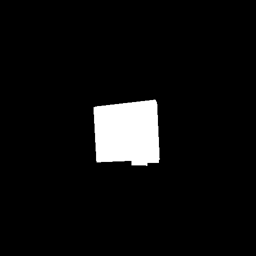}
     \end{subfigure}
     \hfill
     \begin{subfigure}[b]{0.12\textwidth}
         \centering
         \includegraphics[width=\textwidth]{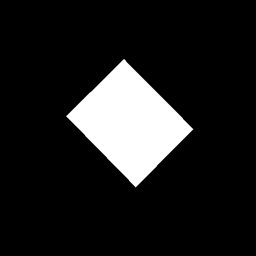}
     \end{subfigure}
     \hfill
     \begin{subfigure}[b]{0.12\textwidth}
         \centering
         \includegraphics[width=\textwidth]{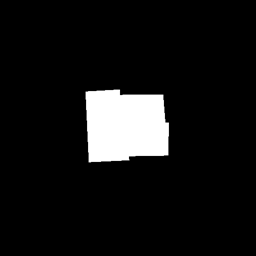}
     \end{subfigure}
     \hfill
     \begin{subfigure}[b]{0.12\textwidth}
         \centering
         \includegraphics[width=\textwidth]{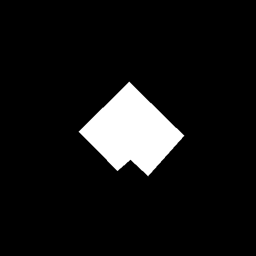}
     \end{subfigure}
     \hfill
     \begin{subfigure}[b]{0.12\textwidth}
         \centering
         \includegraphics[width=\textwidth]{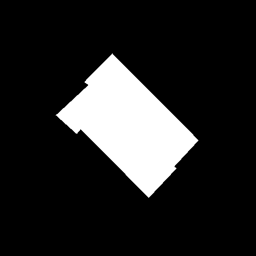}
     \end{subfigure}
     \hfill
     \begin{subfigure}[b]{0.12\textwidth}
         \centering
         \includegraphics[width=\textwidth]{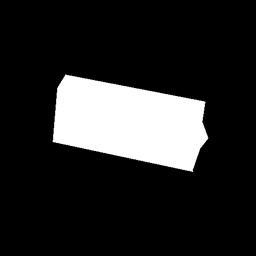}
     \end{subfigure}
    \caption{\textbf{Qualitative results on Vaihingen Buildings~\cite{cramer2010dgpf}.} First row: input image, middle row: our method, bottom row: ground truth.}
    \label{fig:vaihingen_results}
    
\end{figure*}

\begin{table}
  \centering
  \begin{tabular}{@{}lcc@{}}
    \toprule
    Method & mIoU & F1-score \\
    \midrule
    DSAC~\cite{marcos2018learning} & 84.00 & - \\
    TDAC~\cite{hatamizadeh2020end} & 89.16 & - \\
    DARNet~\cite{cheng2019darnet} & 88.24 & - \\
    SegDiff~\cite{amit2021segdiff} & 91.12 & 95.14 \\
    Ours & \textbf{92.50} & \textbf{98.68} \\
    \bottomrule
  \end{tabular}
  \caption{\textbf{Comparison of different methods on Vaihingen buildings}. Results as reported in respective publication. F1-score not reported in all original publications. Our method uses 25 time steps compared to SegDif with 100 time steps.}
  \label{tab:vaihingen}
\end{table}

\subsection{Binary segmentation}\label{sec:binary-segmentation}
The present study provides evidence supporting the effectiveness of our method in binary segmentation settings. Specifically, our results demonstrate that our method can successfully achieve state-of-the-art results in this context.

\paragraph{Data}
We experiment on Vaihingen Buildings~\cite{cramer2010dgpf}, specifically the setup used in \cite{marcos2018learning}, which contains 168 images. These are split between a training set of 100 images and a test set of 68 images. The objective is to label each pixel as either the central building or as background. Multiple buildings can be present in a single image, however, only the building in the center of the image should be labeled. We use the same augmentation strategy as for UAVid but in addition, we also add random (50\%) vertical flips.

\vspace{-0.5em}
\paragraph{Noise function and loss}
We find a simple noise function works best for the binary segmentation setting. Using the same notation as in \Cref{sec:method}, the noise function is defined per pixel as:
\begin{equation}
    \mathbf{s}^{\prime}_{t,\mathbf{\hat{s}}} = \mathbf{\hat{s}}_{t} + \mathbf{z}_{t} \times \frac{t}{T} 
\end{equation}
Where \(\mathbf{z}_{t}\) is sampled from a normal distribution.

For the loss function we use MSE of the predicted added noise as shown in \Cref{eq:mse_loss}.

\paragraph{Architecture} \label{sec:method_architecture}
Our model consists of four modules, a time step head, an image encoder head, a diffused segmentation encoder head and finally the main UNet-like encoder-decoder. The time step head simply converts the time step to a sinusoidal positional embedding, similar to the positional embeddings of \cite{vaswani2017attention}. The image and segmentation heads share a similar structure, consisting of two ResNetBlocks. The output is added together, similar to \cite{amit2021segdiff}. We take inspiration from Efficient U-Net \cite{saharia2022photorealistic} to design our encoder-decoder module. It combines the time step embeddings at multiple stages through the network (in each ResNetBlock). It also utilizes Efficient Attention \cite{shen2021efficient} which is a type of attention with linear complexity. The output of the model is the predicted error (or noise) of the noisy segmentation. The model configuration is shown in \Cref{appendix:architecture}.

\paragraph{Training}
For training, we augment the data with random (50\%) horizontal and vertical flips. In addition, random augmentations to the contrast, saturation and hue. We use AdamW~\cite{loshchilov2017decoupled} as our optimizer with an initial learning rate of \(5 \times 10^{-5}\), decay gamma of \(0.95\) and weight decay of \(1 \times 10^{-3}\). We use gradient clipping at \(1.0\) and train for 70 epochs.

We make a modification to the multi-scale schedule for this task. We convert the linear schedule to a repetitive schedule, which means at each time step, the input is scaled to a different size. This process is shown in \Cref{appendix:multi_scale_schedule}.

\paragraph{Results}\label{sec:experiments_comparison}
The qualitative results of our method on Vaihingen Buildings are presented in \Cref{fig:vaihingen_results}, illustrating its ability to produce segmentations with sharp edges, albeit with some instances of slightly rounded corners. Moreover, \Cref{tab:vaihingen} provides quantitative results on Vaihingen Buildings, where our method achieves the best mIoU.

An analysis of the performance of different model types reveals that diffusion-based models, such as our proposed method and SegDiff, outperform more conventional segmentation approaches. 
Not only does our method outperform SegDiff but it also supports multi-class segmentation.

\section{Discussion} 
Our experiments have demonstrated the feasibility of utilizing diffusion models for multi-class segmentation tasks, yielding results that are comparable to conventional methods while also indicating a higher potential for diffusion models. One possible explanation for this success is the recursive diffusion process, which mimics the training process, and enables the inherent learning of any biases in the noise reduction process. This is the first iteration of our proposed method, and our results suggest that there is ample potential for further research in this area.

Furthermore, it is worth noting that the results presented in this study were obtained with a limited amount of computational resources. The use of more powerful computational resources could potentially yield even better results by enabling the utilization of larger models and longer training times. This is supported by trend seen in \Cref{tab:ablation_training_timesteps}. Additionally, more complex diffusion models and variations thereof could be explored, potentially leading to even greater performance gains. Therefore, further investigation into the use of diffusion models for multi-class segmentation tasks could yield even more advances.

Despite the promising results obtained from utilizing diffusion models for multi-class segmentation, it is important to acknowledge the limitations of this approach. One notable limitation is the higher inference time required for diffusion models, which manifests as a trade-off between accuracy and computational efficiency, as shown in \Cref{tab:ablation_timesteps_inference}. Depending on the specific application, the appropriate balance between accuracy and inference time needs to be considered. However, this limitation is being partly mitigated with the ongoing advancements in hardware capabilities and knowledge distillation techniques \cite{liu2019structured,miles2023mobilevos}. These developments enable the execution of highly accurate models on smaller and more cost-effective devices. Therefore, the use of diffusion models for multi-class segmentation tasks remains a promising option.

Another limitation is the necessity for labeled training data. However, this limitation is a common feature of all supervised segmentation methods. Acquiring sufficient and accurate labeled training data can be a challenging and time-consuming task, especially when dealing with more complex segmentation tasks.
The integration of synthetic data into supervised methods for computer vision has gained momentum as a promising solution to address the challenges posed by limited real-world annotated data \cite{azizi2023synthetic,chen2019learning}. By augmenting training datasets with synthetically generated samples, these methods aim to enhance generalization and robustness, enabling improved performance across various tasks \cite{sandfort2019data}. 

It is interesting to note, the current state-of-the-art for image generation uses diffusion models. Here we use a similar concept for a different purpose. Therefore it would be interesting to explore potential synergies

The use of more powerful hardware can provide the necessary resources to train and run larger and more complex diffusion models, which can potentially lead to further performance gains \cite{kaplan2020scaling,radford2019language,brown2020language}. Additionally, future work can also focus on incorporating recent algorithms and techniques designed to reduce the computational overhead associated with diffusion models \cite{wang2023patch,hang2023efficient}, thereby improving their practicality in real-world applications.

While the utilization of diffusion models for aerial image segmentation tasks has shown promising results, it is worth exploring the applicability of this approach to other segmentation tasks, such as urban street scene segmentation. One such dataset that is commonly used for this task is the Cityscapes \cite{cordts2016cityscapes} dataset, which consists of street scenes from various urban environments. Applying diffusion models to this dataset can potentially provide significant improvements in the accuracy of street scene segmentation, enabling the detection of more fine-grained details and features. However, it is important to consider specific factors that can affect the performance of diffusion models in this different setting. Nonetheless, the potential benefits of utilizing diffusion models for urban street scene segmentation make it an interesting area for further investigation.

Furthermore, a key area for future investigation involves assessing how semantic segmentation diffusion models generalize across different datasets. Analyzing their performance when trained on one dataset and applied to another could reveal insights into their adaptability and robustness.

\section{Conclusion}
In this work, we present \textit{recursive denoising} along with a hierarchical multi-scale diffusion model for semantic segmentation from aerial views. \textit{Recursive denoising} allows for information to propagate through the denoising process. We show our proposed solution yields promising results on UAVid and state-of-the-art results on Vaihingen Buildings. We believe our \textit{recursive denoising} diffusion model is only the first step of a new promising class of segmentation models. Improving aerial segmentation can unlock or improve real-world applications such as delivery, maintenance and, search and rescue, leading to increased efficiency, reduced costs, and improved safety across industries.

\newpage

{\small
\bibliographystyle{ieee_fullname}
\bibliography{bib_new}
}

\newpage

\clearpage
\maketitlesupplementary

\appendix

\begin{figure*}[b]
     \centering
     \begin{subfigure}[b]{0.33\textwidth}
         \centering
         \includegraphics[width=\textwidth]{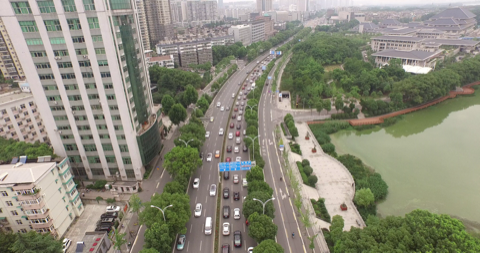}
     \end{subfigure}
     \hfill
     \begin{subfigure}[b]{0.33\textwidth}
         \centering
         \includegraphics[width=\textwidth]{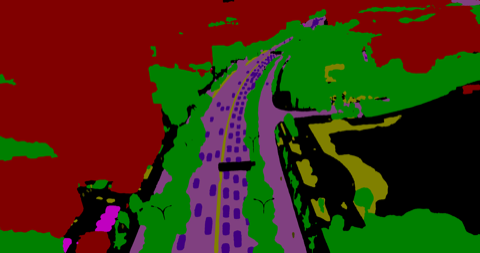}
     \end{subfigure}
     \hfill
     \begin{subfigure}[b]{0.33\textwidth}
         \centering
         \includegraphics[width=\textwidth]{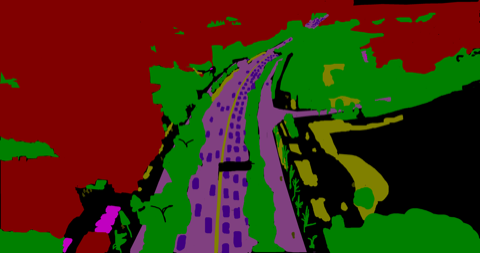}
     \end{subfigure}
     
     \begin{subfigure}[b]{0.33\textwidth}
         \centering
         \includegraphics[width=\textwidth]{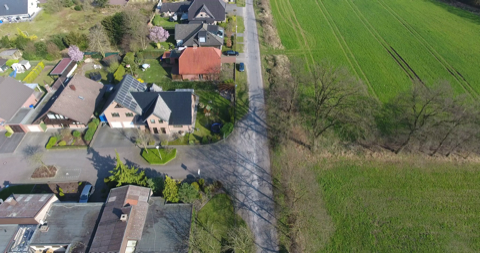}
     \end{subfigure}
     \hfill
     \begin{subfigure}[b]{0.33\textwidth}
         \centering
         \includegraphics[width=\textwidth]{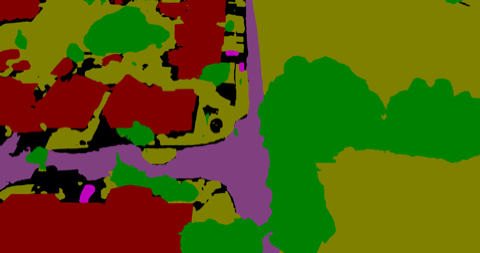}
     \end{subfigure}
     \hfill
     \begin{subfigure}[b]{0.33\textwidth}
         \centering
         \includegraphics[width=\textwidth]{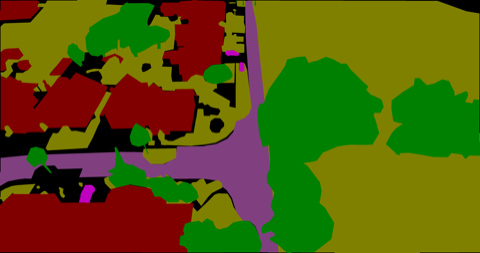}
     \end{subfigure}
     
     \begin{subfigure}[b]{0.33\textwidth}
         \centering
         \includegraphics[width=\textwidth]{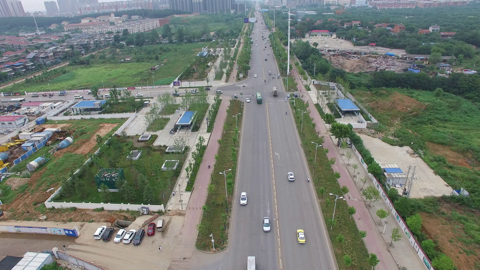}
         \caption{Image}
     \end{subfigure}
     \hfill
     \begin{subfigure}[b]{0.33\textwidth}
         \centering
         \includegraphics[width=\textwidth]{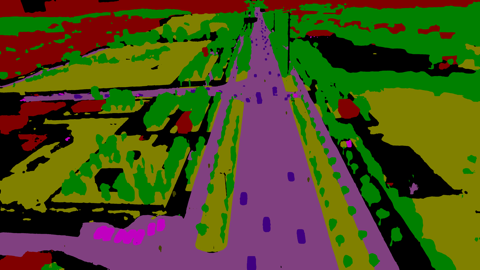}
         \caption{Ours}
     \end{subfigure}
     \hfill
     \begin{subfigure}[b]{0.33\textwidth}
         \centering
         \includegraphics[width=\textwidth]{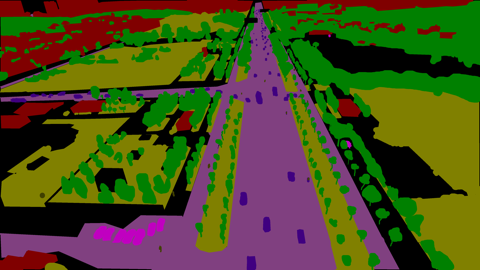}
         \caption{Ground truth}
     \end{subfigure}

    \caption{\textbf{Qualitative results on UAVid~\cite{lyu2020uavid} validation}. Our method can detect small details such as street lights, tree branches and people. The ground truth labels are course and often missing, as can be seen in the first example where the street lights in the center of the image are missing and the tall telephone tower in the bottom example is missing from the ground truth. }
    \label{fig:uavid_results_1}
\end{figure*}

\section{Qualitative experiments}
\label{appendix:qualitative_experiments}
\begin{figure*}[!hp]
     \centering
     \begin{subfigure}[b]{0.33\textwidth}
         \centering
         \includegraphics[width=\textwidth]{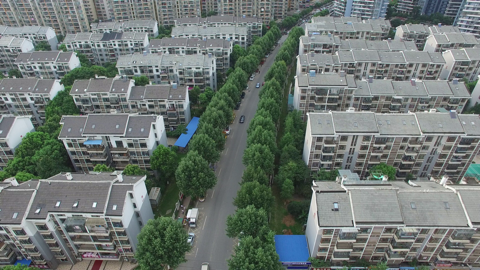}
     \end{subfigure}
     \hfill
     \begin{subfigure}[b]{0.33\textwidth}
         \centering
         \includegraphics[width=\textwidth]{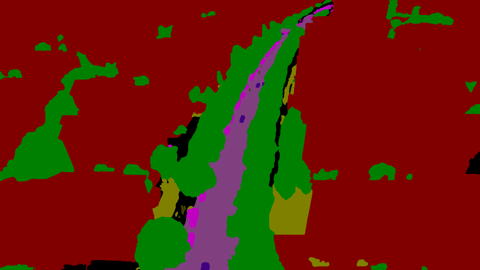}
     \end{subfigure}
     \hfill
     \begin{subfigure}[b]{0.33\textwidth}
         \centering
         \includegraphics[width=\textwidth]{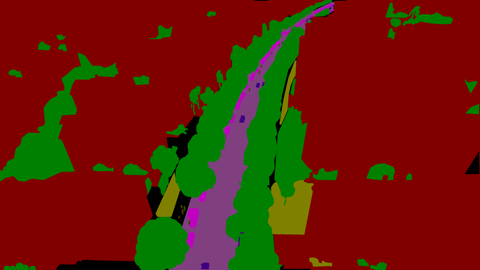}
     \end{subfigure}
     
     \begin{subfigure}[b]{0.33\textwidth}
         \centering
         \includegraphics[width=\textwidth]{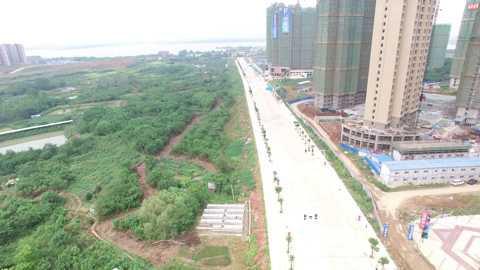}
     \end{subfigure}
     \hfill
     \begin{subfigure}[b]{0.33\textwidth}
         \centering
         \includegraphics[width=\textwidth]{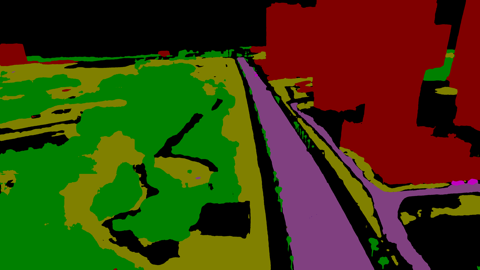}
     \end{subfigure}
     \hfill
     \begin{subfigure}[b]{0.33\textwidth}
         \centering
         \includegraphics[width=\textwidth]{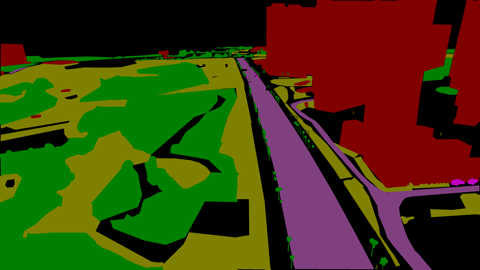}
     \end{subfigure}
     
     \begin{subfigure}[b]{0.33\textwidth}
         \centering
         \includegraphics[width=\textwidth]{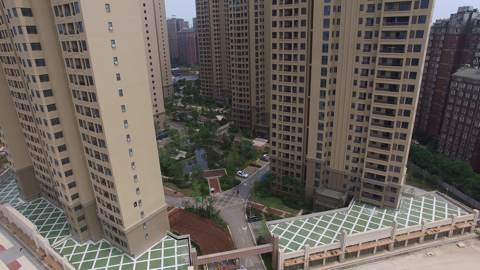}
     \end{subfigure}
     \hfill
     \begin{subfigure}[b]{0.33\textwidth}
         \centering
         \includegraphics[width=\textwidth]{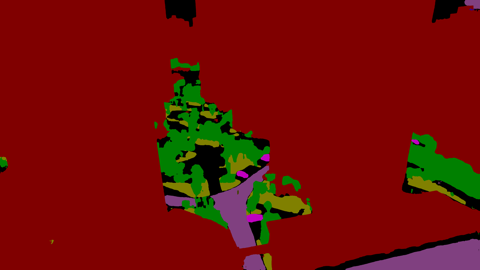}
     \end{subfigure}
     \hfill
     \begin{subfigure}[b]{0.33\textwidth}
         \centering
         \includegraphics[width=\textwidth]{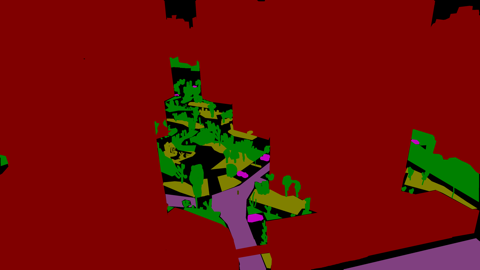}
     \end{subfigure}
     
     \begin{subfigure}[b]{0.33\textwidth}
         \centering
         \includegraphics[width=\textwidth]{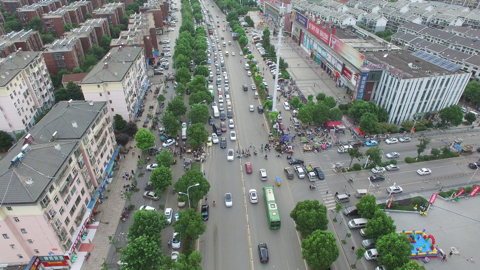}
     \end{subfigure}
     \hfill
     \begin{subfigure}[b]{0.33\textwidth}
         \centering
         \includegraphics[width=\textwidth]{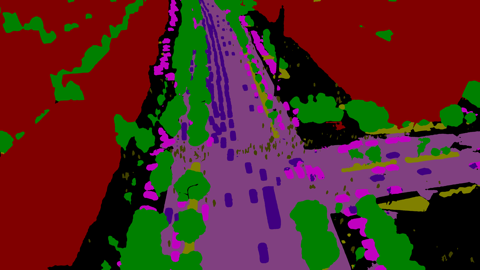}
     \end{subfigure}
     \hfill
     \begin{subfigure}[b]{0.33\textwidth}
         \centering
         \includegraphics[width=\textwidth]{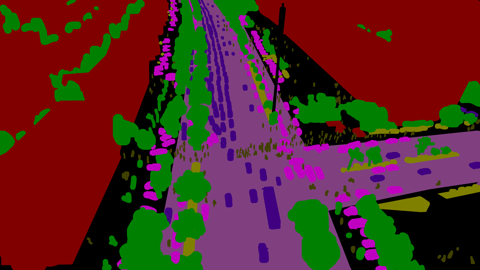}
     \end{subfigure}
     
     \begin{subfigure}[b]{0.33\textwidth}
         \centering
         \includegraphics[width=\textwidth]{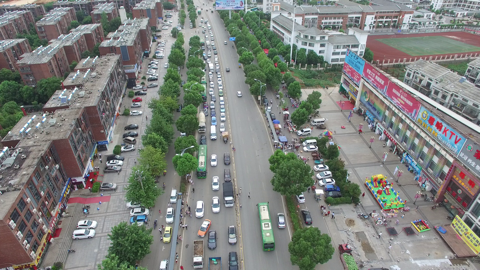}
     \end{subfigure}
     \hfill
     \begin{subfigure}[b]{0.33\textwidth}
         \centering
         \includegraphics[width=\textwidth]{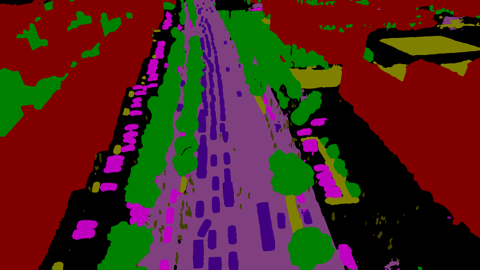}
     \end{subfigure}
     \hfill
     \begin{subfigure}[b]{0.33\textwidth}
         \centering
         \includegraphics[width=\textwidth]{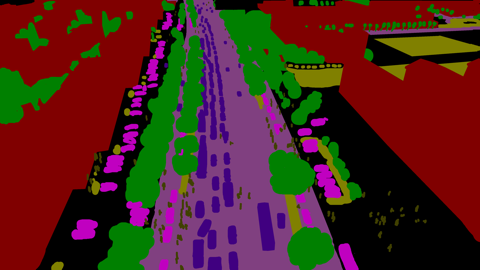}
     \end{subfigure}
     
     \begin{subfigure}[b]{0.33\textwidth}
         \centering
         \includegraphics[width=\textwidth]{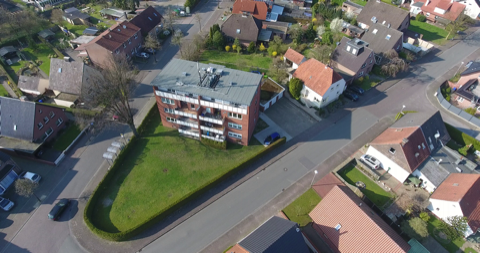}
         \caption{Image}
     \end{subfigure}
     \hfill
     \begin{subfigure}[b]{0.33\textwidth}
         \centering
         \includegraphics[width=\textwidth]{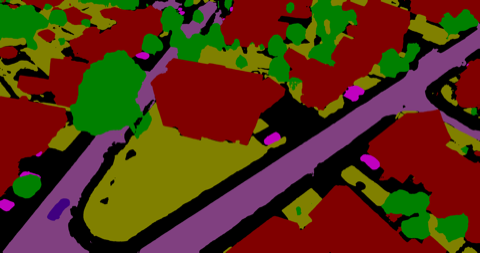}
         \caption{Ours}
     \end{subfigure}
     \hfill
     \begin{subfigure}[b]{0.33\textwidth}
         \centering
         \includegraphics[width=\textwidth]{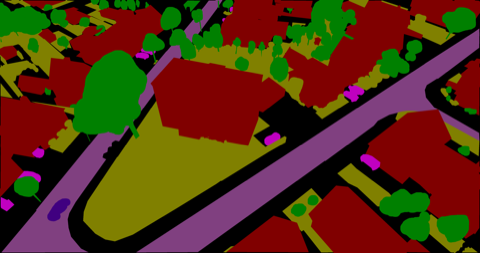}
         \caption{Ground truth}
     \end{subfigure}

    \caption{\textbf{Additional qualitative results on UAVid~\cite{lyu2020uavid} validation}. It is notable how our method can correctly classify static and moving cars given only a single static input image.}
    \label{fig:uavid_results_2}
\end{figure*}

In this section, we provide additional results and qualitative comparisons to further support the effectiveness and robustness of our approach. \Cref{fig:uavid_results_1,fig:uavid_results_2} show qualitative results on the UAVid~\cite{lyu2020uavid} validation dataset, using our multi-class segmentation approach.
Our method shows remarkable ability in segmenting moving cars and static cars, given only a single static input. This indicates our method uses contextual cues found in the surrounding scene to determine if a car is parked or moving. Our method can in some cases produce better segmentation than the ground truth, as can be seen in \Cref{fig:uavid_results_1}.

\begin{figure*}[!hbt]
     \centering
     \begin{subfigure}[b]{0.12\textwidth}
         \centering
         \includegraphics[width=\textwidth]{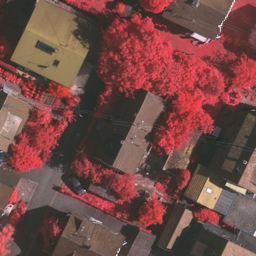}
     \end{subfigure}
     \hfill
     \begin{subfigure}[b]{0.12\textwidth}
         \centering
         \includegraphics[width=\textwidth]{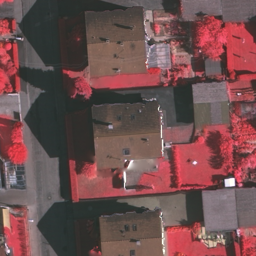}
     \end{subfigure}
     \hfill
     \begin{subfigure}[b]{0.12\textwidth}
         \centering
         \includegraphics[width=\textwidth]{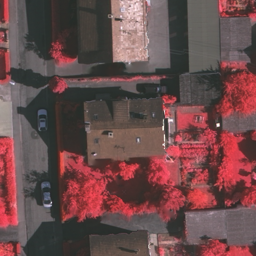}
     \end{subfigure}
     \hfill
     \begin{subfigure}[b]{0.12\textwidth}
         \centering
         \includegraphics[width=\textwidth]{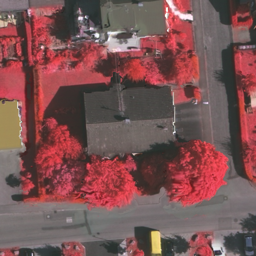}
     \end{subfigure}
     \hfill
     \begin{subfigure}[b]{0.12\textwidth}
         \centering
         \includegraphics[width=\textwidth]{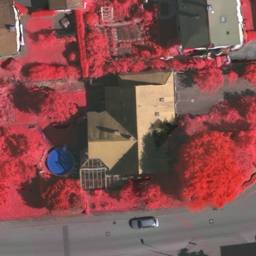}
     \end{subfigure}
     \hfill
     \begin{subfigure}[b]{0.12\textwidth}
         \centering
         \includegraphics[width=\textwidth]{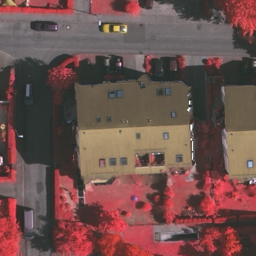}
     \end{subfigure}
     \hfill
     \begin{subfigure}[b]{0.12\textwidth}
         \centering
         \includegraphics[width=\textwidth]{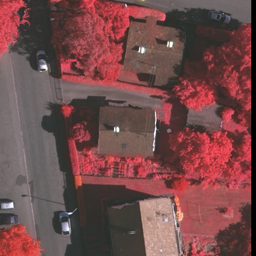}
     \end{subfigure}
     \hfill
     \begin{subfigure}[b]{0.12\textwidth}
         \centering
         \includegraphics[width=\textwidth]{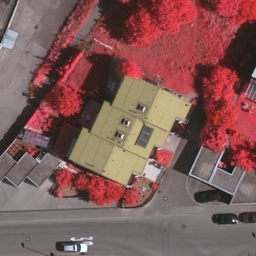}
     \end{subfigure}

     \begin{subfigure}[b]{0.12\textwidth}
         \centering
         \includegraphics[width=\textwidth]{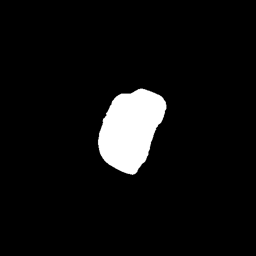}
     \end{subfigure}
     \hfill
     \begin{subfigure}[b]{0.12\textwidth}
         \centering
         \includegraphics[width=\textwidth]{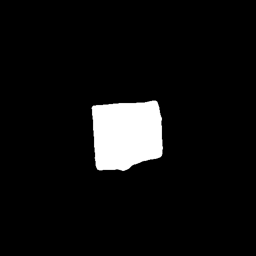}
     \end{subfigure}
     \hfill
     \begin{subfigure}[b]{0.12\textwidth}
         \centering
         \includegraphics[width=\textwidth]{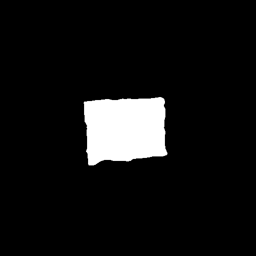}
     \end{subfigure}
     \hfill
     \begin{subfigure}[b]{0.12\textwidth}
         \centering
         \includegraphics[width=\textwidth]{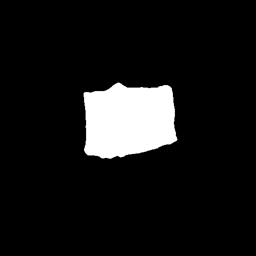}
     \end{subfigure}
     \hfill
     \begin{subfigure}[b]{0.12\textwidth}
         \centering
         \includegraphics[width=\textwidth]{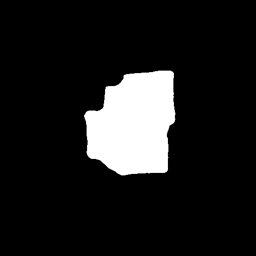}
     \end{subfigure}
     \hfill
     \begin{subfigure}[b]{0.12\textwidth}
         \centering
         \includegraphics[width=\textwidth]{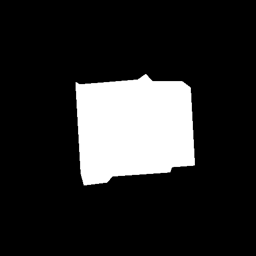}
     \end{subfigure}
     \hfill
     \begin{subfigure}[b]{0.12\textwidth}
         \centering
         \includegraphics[width=\textwidth]{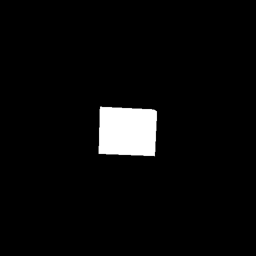}
     \end{subfigure}
     \hfill
     \begin{subfigure}[b]{0.12\textwidth}
         \centering
         \includegraphics[width=\textwidth]{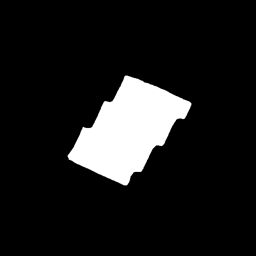}
     \end{subfigure}

     \begin{subfigure}[b]{0.12\textwidth}
         \centering
         \includegraphics[width=\textwidth]{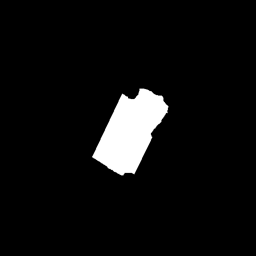}
     \end{subfigure}
     \hfill
     \begin{subfigure}[b]{0.12\textwidth}
         \centering
         \includegraphics[width=\textwidth]{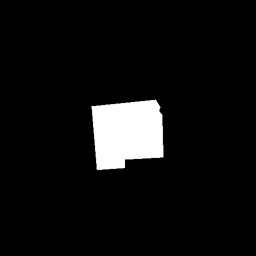}
     \end{subfigure}
     \hfill
     \begin{subfigure}[b]{0.12\textwidth}
         \centering
         \includegraphics[width=\textwidth]{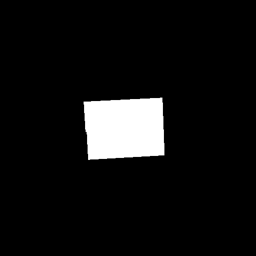}
     \end{subfigure}
     \hfill
     \begin{subfigure}[b]{0.12\textwidth}
         \centering
         \includegraphics[width=\textwidth]{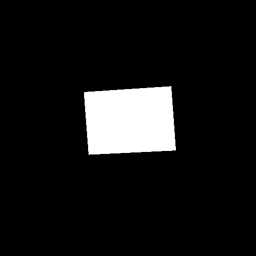}
     \end{subfigure}
     \hfill
     \begin{subfigure}[b]{0.12\textwidth}
         \centering
         \includegraphics[width=\textwidth]{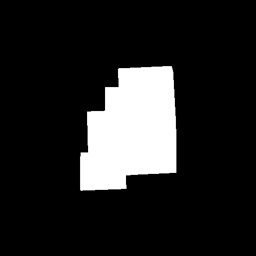}
     \end{subfigure}
     \hfill
     \begin{subfigure}[b]{0.12\textwidth}
         \centering
         \includegraphics[width=\textwidth]{supplementary_figures/vaihingen_buildings/16_seg_gt.png}
     \end{subfigure}
     \hfill
     \begin{subfigure}[b]{0.12\textwidth}
         \centering
         \includegraphics[width=\textwidth]{supplementary_figures/vaihingen_buildings/20_seg_gt.png}
     \end{subfigure}
     \hfill
     \begin{subfigure}[b]{0.12\textwidth}
         \centering
         \includegraphics[width=\textwidth]{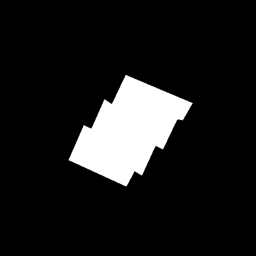}
     \end{subfigure}

    \vspace{-0.20cm}
    \hrulefill
    \vspace{0.18cm}
    
     \begin{subfigure}[b]{0.12\textwidth}
         \centering
         \includegraphics[width=\textwidth]{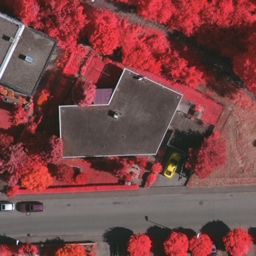}
     \end{subfigure}
     \hfill
     \begin{subfigure}[b]{0.12\textwidth}
         \centering
         \includegraphics[width=\textwidth]{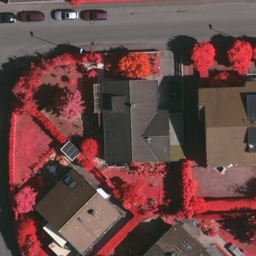}
     \end{subfigure}
     \hfill
     \begin{subfigure}[b]{0.12\textwidth}
         \centering
         \includegraphics[width=\textwidth]{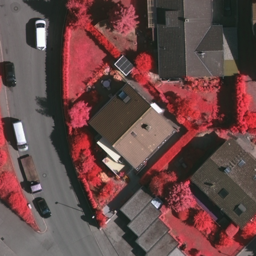}
     \end{subfigure}
     \hfill
     \begin{subfigure}[b]{0.12\textwidth}
         \centering
         \includegraphics[width=\textwidth]{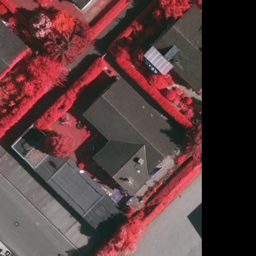}
     \end{subfigure}
     \hfill
     \begin{subfigure}[b]{0.12\textwidth}
         \centering
         \includegraphics[width=\textwidth]{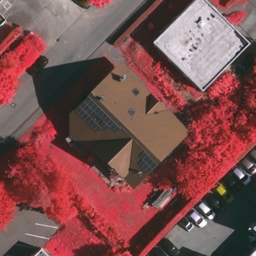}
     \end{subfigure}
     \hfill
     \begin{subfigure}[b]{0.12\textwidth}
         \centering
         \includegraphics[width=\textwidth]{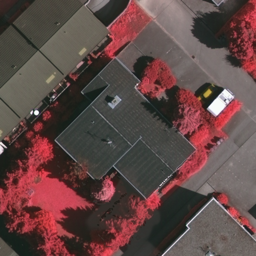}
     \end{subfigure}
     \hfill
     \begin{subfigure}[b]{0.12\textwidth}
         \centering
         \includegraphics[width=\textwidth]{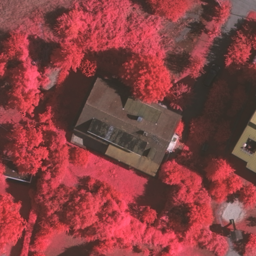}
     \end{subfigure}
     \hfill
     \begin{subfigure}[b]{0.12\textwidth}
         \centering
         \includegraphics[width=\textwidth]{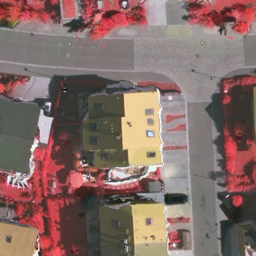}
     \end{subfigure}

     \begin{subfigure}[b]{0.12\textwidth}
         \centering
         \includegraphics[width=\textwidth]{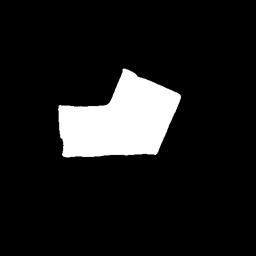}
     \end{subfigure}
     \hfill
     \begin{subfigure}[b]{0.12\textwidth}
         \centering
         \includegraphics[width=\textwidth]{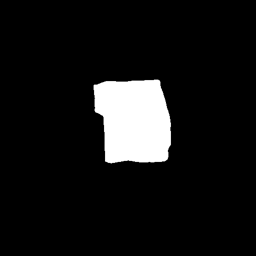}
     \end{subfigure}
     \hfill
     \begin{subfigure}[b]{0.12\textwidth}
         \centering
         \includegraphics[width=\textwidth]{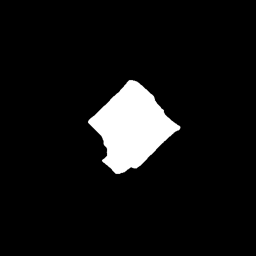}
     \end{subfigure}
     \hfill
     \begin{subfigure}[b]{0.12\textwidth}
         \centering
         \includegraphics[width=\textwidth]{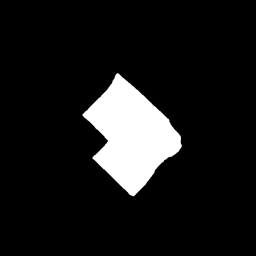}
     \end{subfigure}
     \hfill
     \begin{subfigure}[b]{0.12\textwidth}
         \centering
         \includegraphics[width=\textwidth]{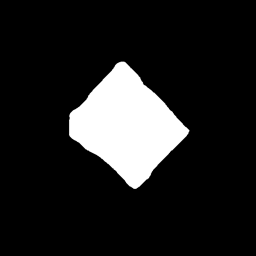}
     \end{subfigure}
     \hfill
     \begin{subfigure}[b]{0.12\textwidth}
         \centering
         \includegraphics[width=\textwidth]{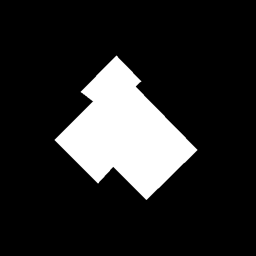}
     \end{subfigure}
     \hfill
     \begin{subfigure}[b]{0.12\textwidth}
         \centering
         \includegraphics[width=\textwidth]{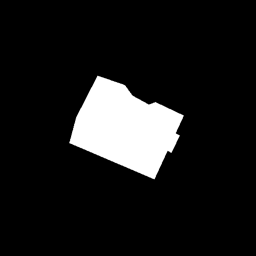}
     \end{subfigure}
     \hfill
     \begin{subfigure}[b]{0.12\textwidth}
         \centering
         \includegraphics[width=\textwidth]{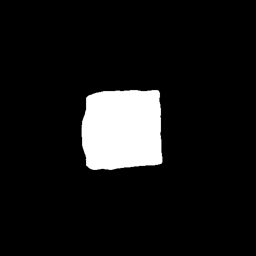}
     \end{subfigure}

     \begin{subfigure}[b]{0.12\textwidth}
         \centering
         \includegraphics[width=\textwidth]{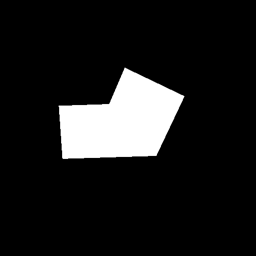}
     \end{subfigure}
     \hfill
     \begin{subfigure}[b]{0.12\textwidth}
         \centering
         \includegraphics[width=\textwidth]{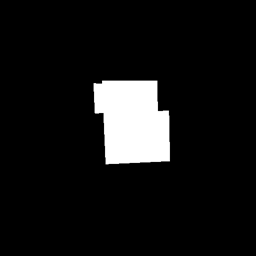}
     \end{subfigure}
     \hfill
     \begin{subfigure}[b]{0.12\textwidth}
         \centering
         \includegraphics[width=\textwidth]{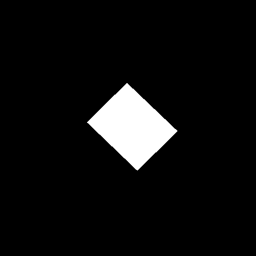}
     \end{subfigure}
     \hfill
     \begin{subfigure}[b]{0.12\textwidth}
         \centering
         \includegraphics[width=\textwidth]{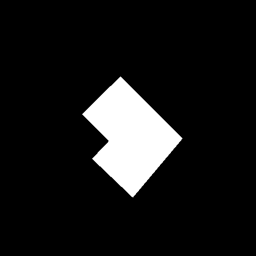}
     \end{subfigure}
     \hfill
     \begin{subfigure}[b]{0.12\textwidth}
         \centering
         \includegraphics[width=\textwidth]{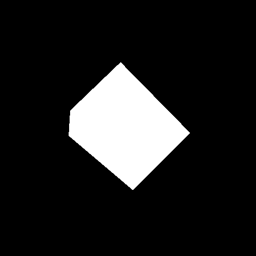}
     \end{subfigure}
     \hfill
     \begin{subfigure}[b]{0.12\textwidth}
         \centering
         \includegraphics[width=\textwidth]{supplementary_figures/vaihingen_buildings/37_seg_gt.png}
     \end{subfigure}
     \hfill
     \begin{subfigure}[b]{0.12\textwidth}
         \centering
         \includegraphics[width=\textwidth]{supplementary_figures/vaihingen_buildings/40_seg_gt.png}
     \end{subfigure}
     \hfill
     \begin{subfigure}[b]{0.12\textwidth}
         \centering
         \includegraphics[width=\textwidth]{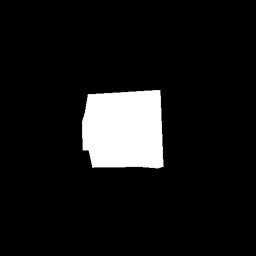}
     \end{subfigure}

    \vspace{-0.20cm}
    \hrulefill
    \vspace{0.18cm}
    
     \begin{subfigure}[b]{0.12\textwidth}
         \centering
         \includegraphics[width=\textwidth]{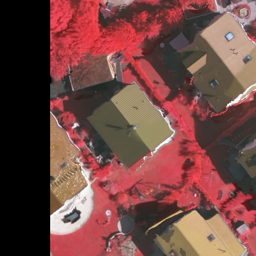}
     \end{subfigure}
     \hfill
     \begin{subfigure}[b]{0.12\textwidth}
         \centering
         \includegraphics[width=\textwidth]{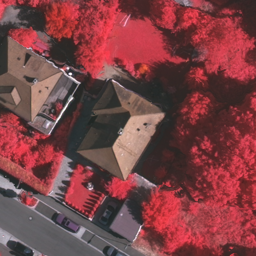}
     \end{subfigure}
     \hfill
     \begin{subfigure}[b]{0.12\textwidth}
         \centering
         \includegraphics[width=\textwidth]{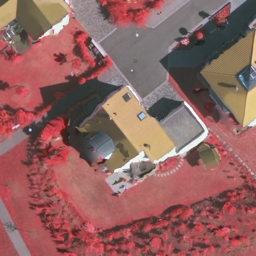}
     \end{subfigure}
     \hfill
     \begin{subfigure}[b]{0.12\textwidth}
         \centering
         \includegraphics[width=\textwidth]{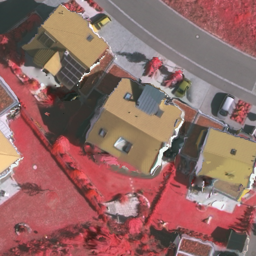}
     \end{subfigure}
     \hfill
     \begin{subfigure}[b]{0.12\textwidth}
         \centering
         \includegraphics[width=\textwidth]{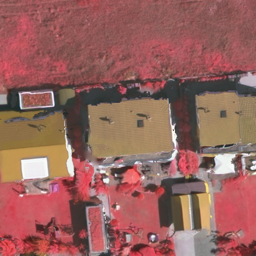}
     \end{subfigure}
     \hfill
     \begin{subfigure}[b]{0.12\textwidth}
         \centering
         \includegraphics[width=\textwidth]{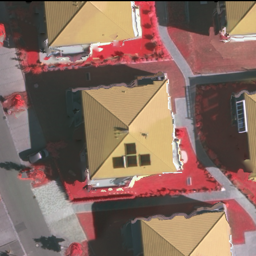}
     \end{subfigure}
     \hfill
     \begin{subfigure}[b]{0.12\textwidth}
         \centering
         \includegraphics[width=\textwidth]{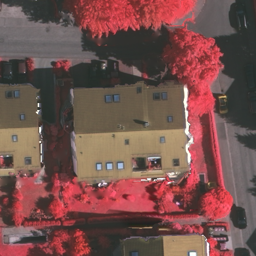}
     \end{subfigure}
     \hfill
     \begin{subfigure}[b]{0.12\textwidth}
         \centering
         \includegraphics[width=\textwidth]{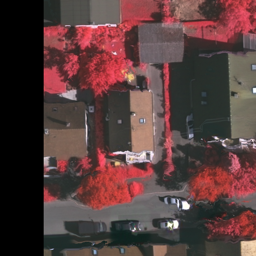}
     \end{subfigure}

     \begin{subfigure}[b]{0.12\textwidth}
         \centering
         \includegraphics[width=\textwidth]{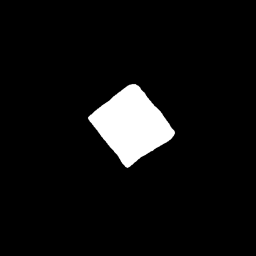}
     \end{subfigure}
     \hfill
     \begin{subfigure}[b]{0.12\textwidth}
         \centering
         \includegraphics[width=\textwidth]{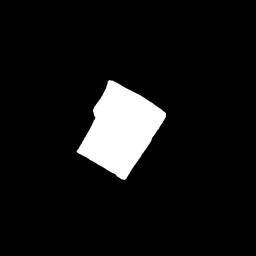}
     \end{subfigure}
     \hfill
     \begin{subfigure}[b]{0.12\textwidth}
         \centering
         \includegraphics[width=\textwidth]{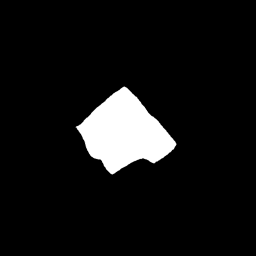}
     \end{subfigure}
     \hfill
     \begin{subfigure}[b]{0.12\textwidth}
         \centering
         \includegraphics[width=\textwidth]{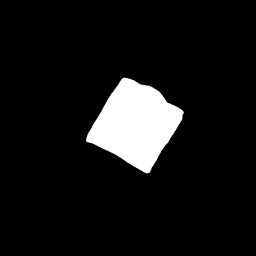}
     \end{subfigure}
     \hfill
     \begin{subfigure}[b]{0.12\textwidth}
         \centering
         \includegraphics[width=\textwidth]{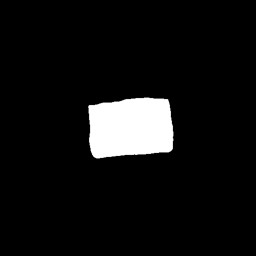}
     \end{subfigure}
     \hfill
     \begin{subfigure}[b]{0.12\textwidth}
         \centering
         \includegraphics[width=\textwidth]{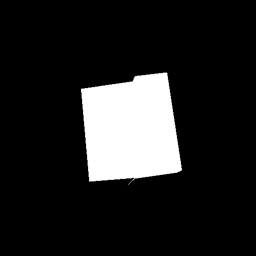}
     \end{subfigure}
     \hfill
     \begin{subfigure}[b]{0.12\textwidth}
         \centering
         \includegraphics[width=\textwidth]{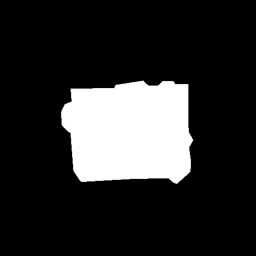}
     \end{subfigure}
     \hfill
     \begin{subfigure}[b]{0.12\textwidth}
         \centering
         \includegraphics[width=\textwidth]{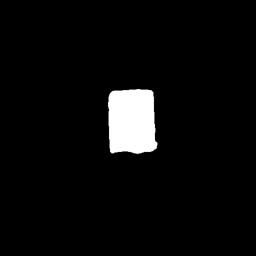}
     \end{subfigure}

     \begin{subfigure}[b]{0.12\textwidth}
         \centering
         \includegraphics[width=\textwidth]{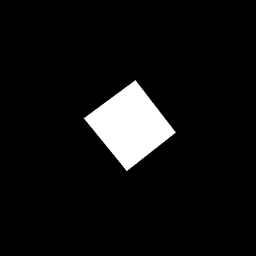}
     \end{subfigure}
     \hfill
     \begin{subfigure}[b]{0.12\textwidth}
         \centering
         \includegraphics[width=\textwidth]{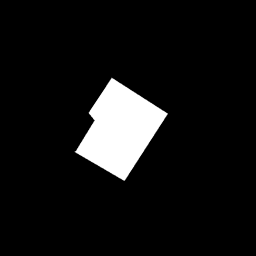}
     \end{subfigure}
     \hfill
     \begin{subfigure}[b]{0.12\textwidth}
         \centering
         \includegraphics[width=\textwidth]{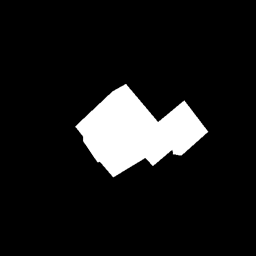}
     \end{subfigure}
     \hfill
     \begin{subfigure}[b]{0.12\textwidth}
         \centering
         \includegraphics[width=\textwidth]{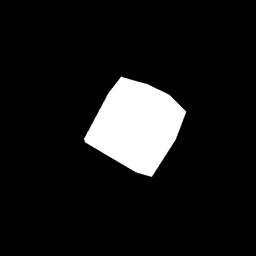}
     \end{subfigure}
     \hfill
     \begin{subfigure}[b]{0.12\textwidth}
         \centering
         \includegraphics[width=\textwidth]{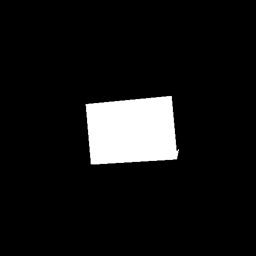}
     \end{subfigure}
     \hfill
     \begin{subfigure}[b]{0.12\textwidth}
         \centering
         \includegraphics[width=\textwidth]{supplementary_figures/vaihingen_buildings/42_seg_gt.png}
     \end{subfigure}
     \hfill
     \begin{subfigure}[b]{0.12\textwidth}
         \centering
         \includegraphics[width=\textwidth]{supplementary_figures/vaihingen_buildings/15_seg_gt.png}
     \end{subfigure}
     \hfill
     \begin{subfigure}[b]{0.12\textwidth}
         \centering
         \includegraphics[width=\textwidth]{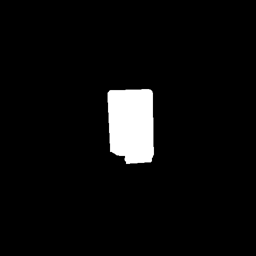}
     \end{subfigure}
    \caption{\textbf{Additional qualitative results on Vaihingen Buildings~\cite{cramer2010dgpf}}. For each row group, top row: input image, middle row: our method, bottom row: ground truth. Our method struggles most when there is a building extension with a different colored roof as can be seen in the bottom row, third from the left.}
    \label{fig:additional_vaihingen_results}
\end{figure*}

\Cref{fig:additional_vaihingen_results} shows additional qualitative results on Vaihingen Buildings~\cite{cramer2010dgpf}. Notably, our method demonstrates a deep understanding of the scene and is able to isolate and segment the primary center building with precision and accuracy, despite the presence of numerous other buildings and structures in the surrounding area.
Images are best viewed on a computer and zoomed-in.

\clearpage
\section{Architecture}
\label{appendix:architecture}

\begin{figure}[ht]
    \centering
    \includegraphics[width=0.9\linewidth]{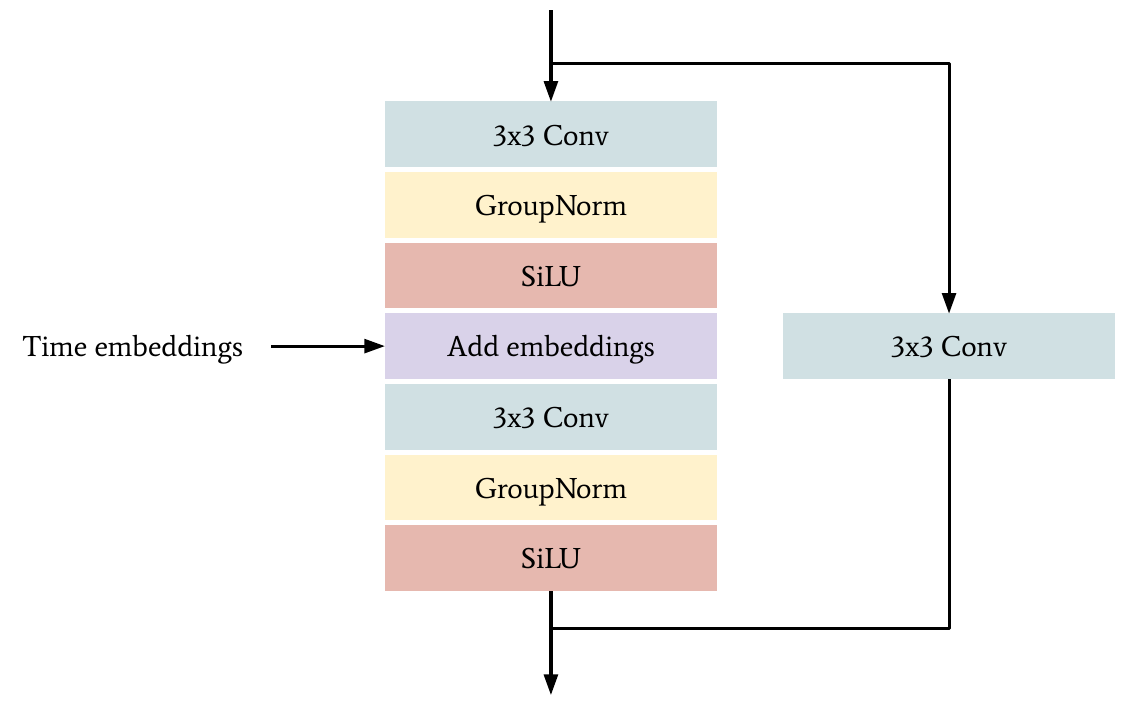}
    \caption{\textbf{Diagram of our ResNetBlock}, consisting of a residual connection \cite{he2016deep}, a core building block for the model.}
    \label{fig:resnetblock}
\end{figure}

\begin{figure}[htb]
    \centering
    \includegraphics[width=0.482\linewidth]{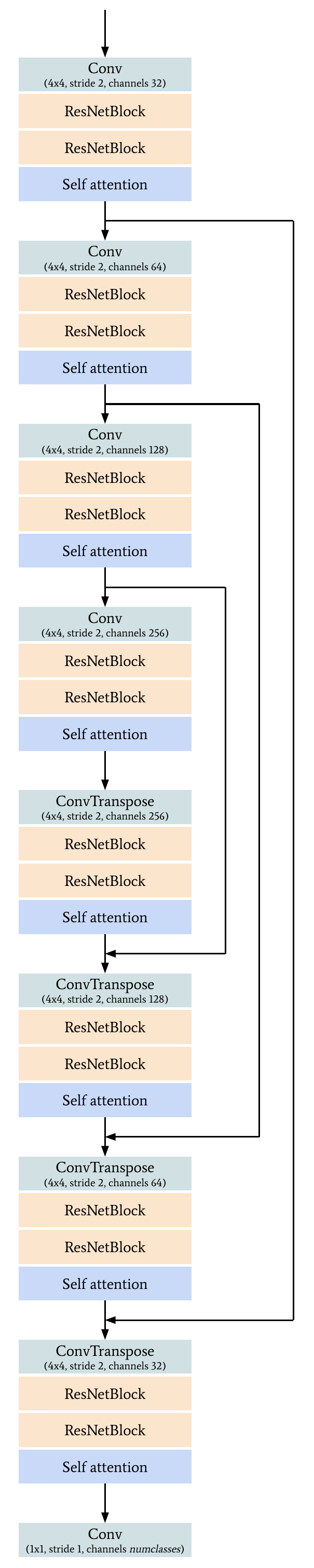}
    \caption{\textbf{Schematic of our encoder-decoder}. The self attention block uses Efficient Attention \cite{shen2021efficient}. The details of the ResNetBlocks are shown in \Cref{fig:resnetblock}. }
    \label{fig:unet}
\end{figure}

Our model is composed of four key modules: a time step head, an image encoder head, a diffused segmentation encoder head, and the primary UNet-like encoder-decoder. To integrate the time step into the model, the time step head transforms it into a sinusoidal positional embedding, inspired by the positional embeddings utilized in Vaswani \etal~\cite{vaswani2017attention}. The image and segmentation heads have the same structure, each including two ResNetBlocks (as shown in \Cref{fig:resnetblock}, but notably without time embeddings). The sum of the outputs, of the image and segmentation heads, are passed to the encoder-decoder. Our encoder-decoder module takes inspiration from Efficient U-Net~\cite{saharia2022photorealistic}. The architecture of our encoder-decoder is shown in \Cref{fig:unet}, which incorporates time step embeddings with each ResNetBlock. Additionally, our model leverages Efficient Attention~\cite{shen2021efficient}, a type of attention mechanism with linear complexity.

\clearpage
\section{Multi-scale schedule variation}
\label{appendix:multi_scale_schedule}
\begin{figure*}[htb!]
    \centering
    \includegraphics[width=1.0\linewidth]{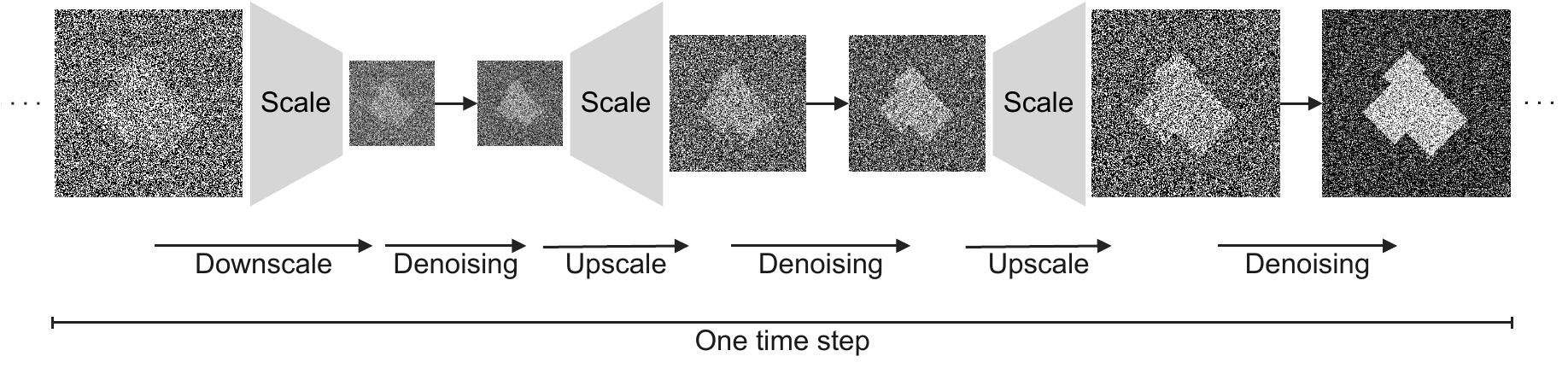}
    \caption{\textbf{Variation on the scaling schedule.} At each time step, the input is resized to its smallest scale and the diffused segmentation is denoised at this lower scale. The denoising process is then repeated iteratively as the segmentation is upscaled until it reaches its original scale.}
    \label{fig:scale_process_loop}
\end{figure*}

For the Vaihingen Buildings dataset we found a modification to the scaling schedule worked better. This modification involves denoising the image at each scale for each time step and can be seen in \Cref{fig:scale_process_loop}. For each time step, the input is downscaled to the smallest scale (if there are multiple scales) and the diffused segmentation is denoised at this smaller scale. Then the segmentation is upscaled and denoised, repeatedly until the original scale is reached. We use bilinear interpolation for both downscaling and upscaling. Training with this scaling schedule is shown in \Cref{alg:multi_scale}.
We found this scaling schedule worked better on Vaihingen Buildings than the linear scale scheduling, which we used for UAVid.

\begin{algorithm}[ht]
\DontPrintSemicolon
  \KwIn{$\mathbf{x} \in \mathbb{R}^{W \times H \times 3}$, RGB image}
  \KwIn{$\mathbf{\bar{s}} \in \mathbb{R}^{W \times H \times \textit{classes}}$, segmentation labels}
  \Parameters{$T \in \mathbb{Z}^{1}$, number of time steps}
  \Parameters{$M \in \mathbb{Z}^{1}$, number of scales}
  $\mathbf{\hat{s}}_{T} \sim \mathcal{N}(\mathbf{0},\mathbf{I})$\;
  \For{$t=T,...,1$}{
   
    \For{$m = M,...,1$}{
        $\text{Resize } \mathbf{\hat{s}}_{t} \text{ to size } (\frac{W}{2^{m-1}} \times \frac{H}{2^{m-1}} \times \text{classes}) $\;
        $\text{Resize } \mathbf{x} \text{ to size } (\frac{W}{2^{m-1}} \times \frac{H}{2^{m-1}} \times 3) $\;
        $\mathbf{z}_{t} \sim \mathcal{N}(\mathbf{0},\mathbf{I})$\;
        $\mathbf{s}^{\prime}_{t} \gets \mathbf{\hat{s}}_{t} + \mathbf{z}_{t} \times \frac{t}{T} $ \tcp*{diffuse}
        $\mathbf{\hat{s}}_{t-1} \gets \mathbf{s}^{\prime}_{t} - \epsilon_{\theta}(\mathbf{s}^{\prime}_{t}, \mathbf{x}, t)$ \tcp*{denoise}
        $l \gets \lVert \epsilon_{\theta} (\mathbf{s}^{\prime}_{t}, \mathbf{x}, t) - (\mathbf{s}^{\prime}_{t} - \mathbf{\bar{s}}) \rVert ^{2}$\;
        $\text{Update } \epsilon_{\theta} \text{ w.r.t. } l$
    }
  }
\caption{Training with hierarchical scales}
\label{alg:multi_scale}
\end{algorithm}

\end{document}